\documentclass{article}

\PassOptionsToPackage{numbers, compress}{natbib}
\usepackage[main, final]{neurips_2026}

\usepackage{todonotes}
\usepackage[utf8]{inputenc} % allow utf-8 input
\usepackage[T1]{fontenc}    % use 8-bit T1 fonts
\usepackage{hyperref}       % hyperlinks
\usepackage{url}            % simple URL typesetting
\usepackage{booktabs}       % professional-quality tables
\usepackage{amsfonts}       % blackboard math symbols
\usepackage{nicefrac}       % compact symbols for 1/2, etc.
\usepackage{microtype}      % microtypography
\usepackage{xcolor}         % colors

\usepackage{graphicx}
\usepackage{subcaption}
\usepackage{amsmath}
\usepackage{siunitx}

\usepackage{enumitem}
\usepackage{multirow}
\usepackage{rotating}

\newcommand{\custompar}[1]{\noindent\textbf{#1:\;}}

\title{Enhancing Tabular Learners with\\ Context-Aware Semantic Embeddings}

\author{%
  Günther Schindler \quad Maximilian Schambach \quad Johannes Höhne \\
  SAP SE \\
  \texttt{\{firstname.lastname\}@sap.com} \\
}

\begin{document}

\maketitle

\begin{abstract}
While modern tabular learners excel at capturing statistical patterns, they frequently operate in a semantic vacuum, treating textual features as discrete symbols, ignoring the rich semantics inherent in feature names or cell entries. 
We propose \textbf{CASE} (\textbf{C}ontext-\textbf{A}ware \textbf{S}emantic \textbf{E}mbeddings), a novel framework that bridges the gap between the semantic understanding of Large Language Models (LLMs) and the statistical capabilities of tabular learners. 
Unlike existing methods that embed rows in isolation, CASE utilizes a contextualization strategy: 
we pre-fill the KV cache of a custom-trained Gemma~3-based Tabular Language Model with a representative sample of rows to establish a persistent anchor of the dataset’s semantics. 
This ensures that generated row embeddings are dynamically contextualized, resolving semantic ambiguities and anchoring representations in domain-specific context. 
Our experiments across several benchmarks (CARTE, TextTab, and TabArena) demonstrate that CASE substantially improves the performance of tabular learners on semantically rich datasets, particularly in low-data regimes.
%\vspace{0.5em} % Adds a tiny, clean vertical gap
\noindent Inference code is available at: \url{https://github.com/SAP-samples/case}.
\end{abstract}

\section{Introduction}

Many real-world tabular prediction tasks include rich textual information such as descriptive column
headers, semantically meaningful categoricals or free-text columns alongside numeric and date-like
features.
The dominance of tabular learners such as Gradient Boosted Decision Trees (GBDTs) and recent in-context learners like TabPFN \cite{tabpfnv2} stem from their ability to find statistical patterns in the feature space.
However, these models operate in a semantic vacuum: they rely entirely on the provided features to infer relationships, often treating categorical values as discrete symbols and lacking native support for free-text columns. 
In contrast, Large Language Models (LLMs) possess vast world knowledge acquired during pre-training, allowing them to recognize concepts, hierarchical relationships, and domain-specific nuances that may not be explicitly found in a single table.
However, one key drawback hinders using LLMs as tabular predictors natively: table serialization and tokenization is comparably inefficient, resulting in very long context-sequences even for relatively small table sizes, limiting how many relevant rows can be provided as context, especially for large datasets. 
In the past, this has severely limited the performance of LLM-based tabular predictors.

The challenge lies in effectively combining the semantic power of LLMs with the statistical reasoning of tabular learners.
Conventional approaches either use heuristic featurization of string or text columns, such as TF-IDF or other n-gram features as used in AutoGluon or the Skrub library, or generate cell- or row-level embeddings in isolation~\cite{contexttab,lefebvre2025knowledgerich}. 
While useful, these methods suffer from a lack of distributional semantic awareness: 
A single row processed in a vacuum lacks the anchor points necessary to resolve ambiguity or interpret values relative to the rest of the dataset.

We introduce \textbf{CASE} (\textbf{C}ontext-\textbf{A}ware \textbf{S}emantic \textbf{E}mbeddings) to bridge this gap. 
First, we train a custom Tabular Language Model (TLM) via continued pretraining of a series of Gemma~3 models~\cite{gemma3} using a column-imputation objective. We apply the loss only to target-column tokens while conditioning on all feature tokens, aligning next-token prediction with tabular prediction.
This yields task-specific representations that capture both the semantic structure of the table and the underlying target distribution.
Next, utilizing this semantic embedder, we employ a ``Context Priming'' strategy: 
By pre-filling the model’s KV cache with a representative sample of rows, including feature as well as target columns from the table's train split, we establish a persistent semantic manifold tailored to the specific dataset. This primed state ensures that when an individual row is embedded, its representation is not merely a translation of its strings, but a contextualized vector positioned relative to the table’s global semantics and predictive task.

We then combine the obtained embeddings with tabular learners to achieve both statistical and semantic grounding:
thus, our approach offers a best-of-both-worlds paradigm: the table-tuned LLM acts as a sophisticated feature engineer with semantic grounding, while the downstream statistical learner performs the heavy lifting of statistical pattern detection and prediction. By projecting the high-dimensional CASE into a lower dimensional space via Principal Component Analysis (PCA), we provide a computationally efficient way to infuse world reasoning into any tabular pipeline.

\begin{figure*}
    \centering
    \includegraphics[scale=0.45]{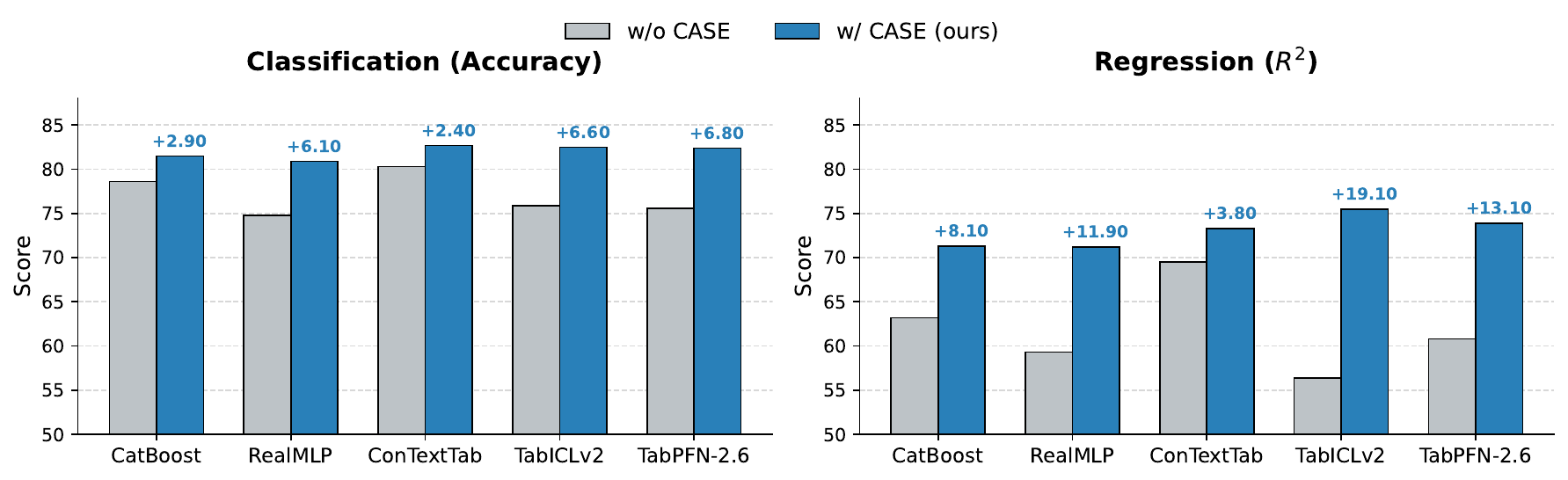}
    \caption{Our proposed \textbf{Context-Aware Semantic Embeddings (CASE)} boost prediction quality of tabular learners on semantically rich benchmarks (CARTE~\cite{CARTE} and TextTab~\cite{texttabbench}).}
    \label{fig:case_comp}
\end{figure*}

%\begin{figure*}
%    \centering
%    \includegraphics[scale=0.5]{figs/elo.pdf}
%    \caption{Elo scores...}
%    \label{fig:elo_scores}
%\end{figure*}

\section{Related Work}

\custompar{Tabular Learning Baselines} Tabular prediction has historically been dominated by GBDTs such as XGBoost, LGBM, or CatBoost~\cite{xgboost, lightgbm, catboost}. While robust, these models lack cross-task transferability and require significant per-dataset hyperparameter optimization and retraining. 
Early deep learning architectures like FT-Transformer~\cite{ft-transformer} and XTab~\cite{xtab} explored transformer-based encoders, but only recent deep learning architectures, such as CARTE, RealMLP, or TabM~\cite{CARTE, realmlp, tabm}, have achieved parity with GBDTs, however, also require re-training and hyperparameter tuning on each table.

\custompar{In-Context Learning (ICL)} Tabular ICL was pioneered by TabPFN \cite{tabpfnv1}, which showed transformers could perform in-context classification on small datasets. This paradigm has evolved from row-level encodings to more scalable cell-based methods like TabPFNv2, TabICL, Mitra, and ConTextTab~\cite{tabpfnv2, tabicl, qu2026tabiclv2, mitra, contexttab}. While these models excel at structural pattern matching, they often treat values as abstract tokens, overlooking the latent semantic richness inherent in the data.
While ConTextTab uses semantic embeddings natively, these embeddings are obtained cell-wise using a small sentence embedder and are subsequently contextualized by the ICL model. The model itself, however, is not trained on semantic prediction tasks like LLMs but rather on in-context learning target prediction which may limit its ability for advanced semantic reasoning  .% is unlikely to foster higher-level semantic reasoning.

\custompar{LLMs and Table Semantics} 
Models like TabLLM, LIFT, TabuLa, and TabGemma~\cite{tabllm, lift, tabula8b, tabgemma} leverage the world knowledge of LLMs to add semantics to tabular predictions.
%To bridge the gap between statistics and semantics, models like TabLLM, LIFT, and TabuLa~\cite{tabllm, lift, tabula8b} leverage the world knowledge of LLMs. 
These methods excel in low-data regimes where semantic priors compensate for sparse signals, though they often struggle with numeric-heavy datasets and longer context sizes.
Fundamentally, due to the comparably inefficient tokenization, even recent large-context window LLMs may not fit a full table into their context, while table-native ICLs now scale to 10's or 100's of thousand context rows.
Other works have also explored LLM- or knowledge graph-based enrichment of conventional tabular learners~\cite{CARTE,lefebvre2025knowledgerich} or experimented with creating target-aware embeddings using per-table tuned text encoders~\cite{tabstar}.

% We introduce CASE, a framework for enhancing arbitrary tabular learners with context-aware semantic embeddings. The semantic embeddings are generated by TabGemma, a specialized adaptation of the Gemma 3 decoder-only architecture. TabGemma is pre-trained on the T4 corpus using a Tabular Imputation Objective which align embeddings towards tabular predictions tasks. To establish the semantic context of the table using TabGemma’s limited context window of 128k tokens, we employ a Context Priming strategy. We apply Principal Component Analysis (PCA) to reduce the raw embedding into compact components.

\section{Methodology}
We introduce CASE, a framework that enhances tabular learners with context-aware semantic embeddings. At its core, we develop a Tabular Language Model (TLM) derived from the Gemma 3 decoder-only family, pretrained via a custom table serialization and an in-context target prediction objective. By leveraging this TLM, we extract row representations that are contextualized, i.e.\ conditioned on the table's specific semantic distribution and its predictive task. These semantic embeddings are subsequently compressed and fused with the native features to be used together with any downstream tabular learner, combining the rich world knowledge of language models with the statistical precision and predictive capabilities of specialized tabular architectures.

\begin{figure*}
    \centering
    \includegraphics[width=\linewidth]{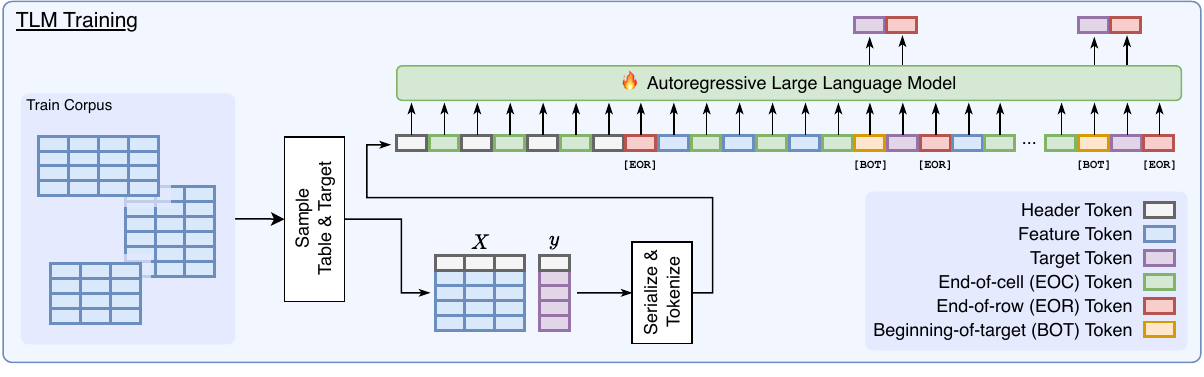}
    \caption{Illustration of our proposed Tabular Language Model (TLM) architecture with table serialization and target-imputation objective during training.}
    \label{fig:embedding_model}
\end{figure*}

\subsection{Tabular Language Model Pretraining}
To extract context-aware row representations, we perform continued pretraining on a diverse suite of tabular prediction tasks synthesized from a large-scale collection of real-world datasets. Following the paradigm of TabuLa~\cite{tabula8b}, we frame tabular data as a conditional sequence modeling problem. 
Tables are serialized into token sequences where a long-context LLM is trained to predict a designated target column, causally conditioned on the preceding feature tokens and the broader dataset context. 
By casting tabular prediction as an autoregressive generative process, the model learns to synthesize its internal world knowledge with observed statistical patterns. 
This results in embeddings that are inherently task-aligned and grounded in the table’s specific semantic. An overview of this pretraining architecture is illustrated in Figure~\ref{fig:embedding_model}.

\custompar{Table Serialization and Tokenization} 
We serialize each table row into a linear token sequence. 
Every cell is first cast to a canonical string, tokenized, and concatenated in column order. 
Cells are separated by a dedicated cell-separator token, and each row is terminated with an end-of-row token.

That is, for a table consisting of $F$ features $X$ and a target $y$ with $N$ rows, we define a serialization function $\mathcal{S}(\cdot)$ that transforms a table into a continuous token sequence, beginning with the header sequence $\mathbf{H}$, which establishes the semantic identity of the columns. The header sequence is obtained by concatenating the tokenized column headers $c_k$, adding end-of-cell (EOC) tokens in between and finalizing the sequence with an end-of-row (EOR) token:
\begin{equation}
    \mathbf{H} = c_1 \parallel \texttt{[EOC]} \parallel c_2 \parallel \texttt{[EOC]} \parallel \cdots \parallel c_{F+1} \parallel \texttt{[EOC]} \parallel \texttt{[EOR]} \,,
\end{equation}
where $\Vert$ denotes concatenation along the sequence axis.
Following the header, each row $r_i$ of the table is serialized by concatenating the tokenized feature values $v_{i,k}$ and the final target value $y_i$, separated by the Beginning-of-Target (BOT) marker:
\begin{equation}
    \mathcal{S}(r_i) =  v_{i,1} \parallel \texttt{[EOC]} \parallel \cdots \parallel v_{i,F}  \parallel \texttt{[EOC]} \parallel \texttt{[BOT]} \parallel y_i  \parallel \texttt{[EOR]} \,.
\end{equation}

The total input sequence $\mathbf{S}$ obtained from the table is the concatenation of the header sequence and all row sequences:
\begin{equation}
    \mathbf{S} = \mathbf{H} \parallel \mathcal{S}(r_1) \parallel \dots \parallel \mathcal{S}(r_N) \,.
\end{equation}
Throughout, numerical targets are normalized into a scientific notation (for example, 3141.592 becomes +3.1416e+03) and other data types are natively handled by the LLM's tokenizer.

\custompar{Training Objective} 
We initialize our model from the pretrained Gemma 3 checkpoint (without instruction-tuning), which supports a 128k-token context window, and continue pretraining the model on a tabular imputation objective.
Given an input table with feature-target split, our approach is trained using a supervised objective where the next-token prediction loss is computed exclusively over target tokens. Let $\mathcal{I}_{\textrm{target}}$ be the set of indices in $\mathbf{S}$ corresponding to the tokens of the target values $\{y_1, \dots, y_N\}$. The training objective minimizes the target-masked negative log-likelihood,
\begin{equation}
    \mathcal{L}(\theta) = - \sum_{t \in \mathcal{I}_{\textrm{target}}} \log P(s_t \mid s_{<t}; \theta) \,,
\end{equation}
with respect to the model weights $\theta$.
This objective forces the model to predict target tokens from the \texttt{[BOT]} embedding conditioned on its preceding context.
Note that, due to the autoregressive nature of the used LLM and the use of causal attention masking, every target cell is used as a training sample and the approach does not require a fixed context-query split unlike table-native approaches such as TabPFN -- practically increasing the effective batch size as well as conditioning the model on different effective context lengths within a single training step.

\custompar{Training Data} 
We train our approach on the large-scale real-world T4 table corpus~\cite{tabula8b}. 
Similar to previous works~\cite{contexttab,tabdpt,GeneralizationCanEmerge}, we generate tabular prediction tasks from these unlabeled datasets: 
For each training step, we draw a table from the corpus, uniformly sample 256 rows, and designate one column as the prediction target while using the remaining columns as conditioning features.

\custompar{Training Setting}
We train different base model sizes (270M, 1B, 4B, and 12B parameters) for \num{8500} steps with a batch size of 128, corresponding to roughly 280 million rows in total. We uses Adam with a learning rate of $10^{-6}$. 
We do not apply dropout or weight decay. 
To balance throughput and context utilization, we cap inputs at 16k tokens during training and truncate longer sequences.

\begin{figure*}
    \centering
    \includegraphics[width=\linewidth]{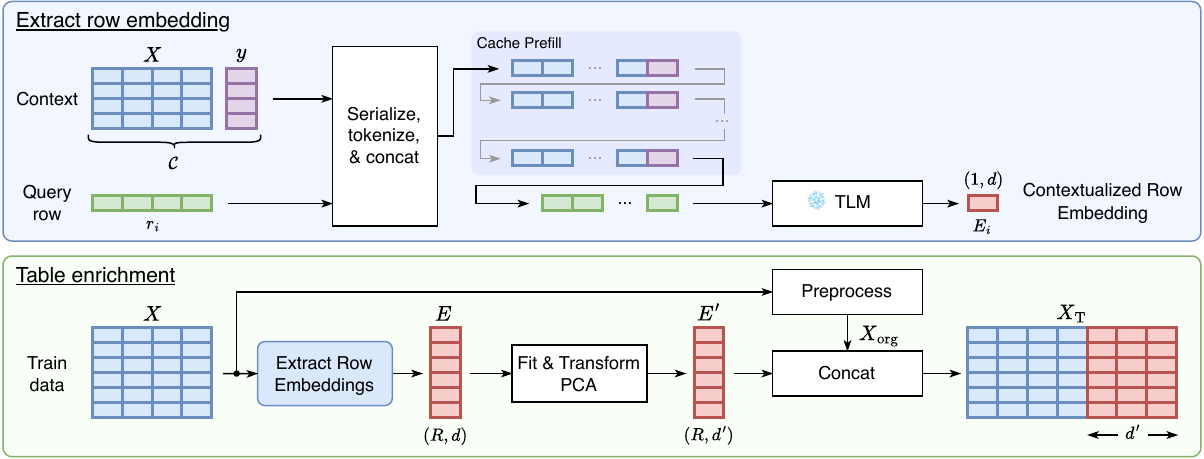}
    \caption{Overview of our proposed context-aware semantic embedding-based table enrichment pipeline. We pre-fill the KV cache of a table-native pretrained LLM using a context subsample of the training data. Rows are embedded using the contextualized \texttt{[BOT]} token of the sequence. For enrichment of the feature matrix $X$, individual rows are embedded, PCA-transformed and appended to the (potentially preprocessed) feature matrix.}
    \label{fig:architecture}
\end{figure*}

\subsection{Table Enrichment via Context-Aware Semantic Embeddings}
Given a table corresponding to a predictive task with features $X$ and target $y$, our goal is to perform semantic enrichment of $X$ as part of a tabular data featurization pipeline usable with any tabular predictor.
To this end, we leverage the contextualized row-wise embeddings obtained from our pretrained TLM.
An overview of the proposed approach is given in Figure~\ref{fig:architecture}.

\custompar{Context Priming and Contextualized Encoding}
 As opposed to creating isolated, row-wise embeddings for each row of the table, as investigated in previous works~\cite{grinsztajn2023vectorizing,lefebvre2025knowledgerich}, our goal is to establish a global semantic context for the row embeddings, grounding each embedding within the table's context and predictive task.
 To this end we employ a context priming strategy: 

First, a random subset of rows is sampled as context $\mathcal{C}$, containing both features $X$ and targets $y$ from the task's train split, serialized and passed through the previously trained TLM. 
Second, the resulting KV cache is retained as a permanent prefix, distilling the table's semantic distribution for all subsequent row encodings.
Finally, each target row $r_i$ is passed through the model, reusing the pre-filled KV cache. 
We extract the row embedding $\mathbf{e}_i$ from the final hidden layer at the final \texttt{[BOT]} token position. 
This embedding captures the integrated semantic summary of the record before the prediction would be made by the LLM, conditioning the embedding both on the global semantic context as well as the underlying predictive task.
This is crucially different to existing approaches, which perform either cell- or row-wise, uncontextualized and target-agnostic embeddings.

\custompar{Table Enrichment}
In order to enrich the input feature matrix $X$ with the obtained embeddings, the goal is to append the row embeddings along the feature axis as additional numerical features.
However, the obtained row embeddings are high-dimensional, with embedding dimension up to $d=4096$ for a Gemma 3~12B-based TLM.
To reduce the dimensionality, preventing memory or performance issues when appended to the feature matrix for the tabular learners, we apply a Principal Component Analysis (PCA) to the set of embeddings $\{\mathbf{e}_1, \dots, \mathbf{e}_n\}$, reducing them to $d'$ components. 
The final enriched feature matrix is:
$ X_{\textrm{T}} = [X_{\textrm{org}} \parallel \operatorname{PCA}(\mathbf{E})] $ ,
where $X_{\textrm{org}}$ are the original, potentially preprocessed features, depending on the tabular learner used.
To prevent data leakage, PCA components are fitted solely on the training distribution. These frozen components are subsequently applied to test-row embeddings during the inference phase.

\begin{table*}[t]\centering\footnotesize
\caption{Evaluation results on the investigated benchmarks, reporting rank (Rk), accuracy (Acc) and (soft-clipped) R$^{\textrm{2}}$ in percent for classification and regression.}
\label{tab:tabular-benchmarks}
\begin{tabular}{lrrrrrrrrrrrr}
\toprule
 & \multicolumn{3}{c}{\textbf{All}} & \multicolumn{3}{c}{\textbf{CARTE}} & \multicolumn{3}{c}{\textbf{TextTab}} & \multicolumn{3}{c}{\textbf{TabArena}} \\
  \cmidrule(lr){2-4} \cmidrule(lr){5-7} \cmidrule(lr){8-10} \cmidrule(lr){11-13}
Model & \multicolumn{1}{c}{\textbf{Rk}} & \multicolumn{1}{c}{\textbf{Acc}} & \multicolumn{1}{c}{\textbf{R$^\textrm{2}$}}  & \multicolumn{1}{c}{\textbf{Rk}} & \multicolumn{1}{c}{\textbf{Acc}} & \multicolumn{1}{c}{\textbf{R$^\textrm{2}$}} & \multicolumn{1}{c}{\textbf{Rk}}  & \multicolumn{1}{c}{\textbf{Acc}} & \multicolumn{1}{c}{\textbf{R$^\textrm{2}$}}  & \multicolumn{1}{c}{\textbf{Rk}} & \multicolumn{1}{c}{\textbf{Acc}} & \multicolumn{1}{c}{\textbf{R$^\textrm{2}$}}  \\
\midrule
\bfseries TabICLv2 w/ CASE \hspace{-1em} & \bfseries 2.7 & \bfseries 86.5 & \bfseries 76.1 & \bfseries 1.3 & \bfseries 81.0 & \bfseries 77.5 & \bfseries 2.9 & \bfseries 85.7 &  68.2 & 4.0 & 88.3 & 78.5 \\
AutoGluon & 3.1 & 85.6 & 73.9 & 2.6 & 78.8 & 73.7 & 3.7 & 83.5 & 67.5 & 3.4  & 88.1 & 79.9 \\
ConTextTab & 3.8 & 85.1 & 71.2 & 3.2 & 77.1 & 72.4 & 3.9 & 84.4 & 58.8 & 4.4 & 87.6 & 77.8 \\
RealMLP  & 4.0 & 84.8 & 70.8 & 4.5 & 74.6 & 68.6 & 4.0 & 82.0 & \bfseries 68.3 & 3.6 & 88.4 & 79.7 \\
CatBoost  & 4.2 & 85.2 & 69.9 & 4.7 & 76.3 & 68.3 & 3.6 & 83.7 & 65.4 & 3.8 & 88.2 & 79.0 \\
TabPFN-2.6 & 4.3 & 84.2 & 64.8 & 6.7 & 70.7 & 59.9 & 4.6 & 81.5 & 63.8 & \bfseries 1.8 & \bfseries 88.7 & \bfseries 80.5 \\
TabICLv2 & 4.5 & 84.3 & 61.1 & 6.9 & 70.5 & 55.3 & 4.5 & 82.5 & 60.2 & 2.1 & \bfseries 88.7 & 79.9 \\
XGBoost  & 5.3 & 84.2 & 69.1 & 5.7 & 73.5 & 66.9 & 5.1 & 81.4 & 65.4 & 5.1 & 87.9 & 78.8 \\
Naive & 8.5 & 70.1 & -3.5 & 9.0 & 53.0 & -1.7 & 8.6 & 70.4 & -5.5 & 8.1 & 75.0 & -7.3 \\
%Gemma 3 12B & \NAN & \NAN & 4.6 & -98.4 & \NAN & \NAN & \NAN & \NAN \\
\bottomrule
\end{tabular}
\end{table*}

\section{Experiments and Results}

% Moved to the previous section for TabGemma training
% \custompar{Setup and Training}
% TODO: add details on which model and training data used, number of epochs, etc.\todo{TODO}

\custompar{Evaluation} 
We evaluate our approach across three distinct benchmark suites: CARTE~\cite{CARTE}, TextTab~\cite{texttabbench}, and the single-fold variant of TabArena~\cite{tabarena}. 
Our focus lies on CARTE and TextTab as they are specifically designed to emphasize semantic relationships in tabular data, aligning with the core motivation of our work. 
Conversely, TabArena serves as a numerics-heavy baseline, included to test the robustness of our approach in environments where statistical signals dominate over semantic ones. 
All benchmarks cover both classification and regression tasks.
Throughout, we report mean accuracy for classification and mean (soft-clipped) R\textsuperscript{2} for regression tasks, respectively, following previous works~\cite{contexttab}.

\custompar{Baselines} We compare against a diverse set of state-of-the-art baselines, covering conventional per-dataset trained and HPO-tuned ones (XGBoost, CatBoost, and RealMLP), tabular in-context learners (TabPFN-2.6, TabICLv2, and ConTextTab), and the AutoGluon framework, using the ``best`` preset with a 4\,h time limit~\cite{autogluon}.
In particular, note that we combine the table-tuned baselines (GBDTs and RealMLP) with the AutoGluon feature preprocessor which includes some text feature handling via n-gram features for free text or high-cardinal features.
We perform ablations with other forms of text featurization in the subsequent sections.
%Also note that, despite much effort spent, we were not able to produce results using the public checkpoint and code of TabuLa~\cite{tabula8b}, the most recent TLM.
Additional details on the used baselines are provided in Appendix \ref{app:baselines}.

% \begin{itemize}
% \item Tabular ICL: State-of-the-art in-context learners including \textbf{ConTextTab} and \textbf{TabICL}.
% \item Language-model ICL: \textbf{TabGemma} used as a standalone predictor to establish a few-shot baseline.
% \item AutoML: The \textbf{AutoGluon} \cite{autogluon} framework, representing high-performance automated statistical modeling.
% \item Baseline: A naive predictor to establish the performance floor.
% \end{itemize}

\custompar{Default setup} 
For our primary evaluation of CASE, we utilize the custom Gemma 3 12B-based TLM with a context priming window of $k\,{=}\,128$ rows (with a maximum of 32k tokens) and a PCA-reduced latent space of $d'=32$ for the feature fusion.

\subsection{Tabular Learners benefit from semantic enrichment via CASE}
We evaluate the efficacy of the CASE framework by enriching the input features for five prominent tabular learning architectures, spanning conventional as well as recent deep learning baselines, both per-table tuned as well as in-context learning approaches. 
Namely, we evaluate CatBoost, RealMLP, TabPFN-2.6, TabICLv2, and ConTextTab, representing best-in-class models in each category.

%We compare these baselines in their default preprocessing configurations (using the native feature handling in the case of TabPFN, TabICL, and ConTextTab, and using AutoGluon-based feature preprocessing of textual features for CatBoost and RealMLP) against their CASE-enriched counterparts. 
The results on the semantics-heavy CARTE and TextTab benchmarks, covering a total of 20 classification and 51 regression tasks, are shown in Figure~\ref{fig:case_comp}.
% We observe that CASE is universally increasing the performance of each investigated predictor, boosting mean accuracy by up to \SI{6.8}{\percent}, even increasing the performance of the latest state-of-the-art, the semantic-native ConTextTab, by \SI{2.4}{\percent}.
% CASE effectively closes the gap for models that otherwise struggle with semantically rich data, such as TabPFN-2.6 or TabICLv2, setting a new state-of-the art.
% The performance increase is even more impressive on regression taasks, improving R\textsuperscript{2} by up \num{19.1} percentage points.
% Even ConTextTab, which internally uses cell-wise langauge model embeddings, sees a significant bump in performance when combined with CASE, indicating the additional benefit of contextualized semantic embeddings, which we will ablate in more detail later.

% Overall, CASE is highly effective across all investigated tabular prediction models, including GBDT, recent deep learning approaches, as well as state-of-the-art tabular in-context learners.

\custompar{Quantitative Gains} 
We observe that CASE yields a large accuracy uplift ranging from \textbf{2.9\% to 6.80\%} in classification tasks, and a remarkable R\textsuperscript{2} gain between \textbf{3.8\% and 19.1\%} in regression tasks. These results suggest that our novel approach towards semantic embeddings successfully capture latent dependencies that are inaccessible to all investigated models, regardless of their architecture.
%old text
%As illustrated, the integration of CASE leads to substantial improvements in predictive power. We observe a significant accuracy uplift ranging from \textbf{2.9\% to 6.80\%} in classification tasks, and a remarkable $R^2$ gain between \textbf{3.8\% and 19.1\%} in regression. These results suggest that TabGemma’s semantic embeddings successfully capture latent dependencies that are inaccessible to models relying solely on raw statistical features.

\custompar{Synergy with Semantic Models} 
Notably, even ConTextTab -- which natively incorporates cell-wise text embeddings using a pretrained Sentence Transformer -- benefits significantly from CASE. This indicates that our target-aware, contextualized embeddings provide a more nuanced representation of tabular relationships that complement general-purpose sentence embedding used in current models.
%Notably, even \textit{ConTextTab}—which natively incorporates semantic priors via Sentence Transformer pre-training—benefits significantly from CASE. This indicates that TabGemma’s target-aware, contextualized embeddings provide a more nuanced representation of tabular relationships than general-purpose sentence encoders.

\custompar{Revitalizing ICL Models} A key finding is that CASE successfully boost performance of recent tabular ICL models without native text handling like TabICLv2 and TabPFN-2.6. While these models excel at numerically dominated benchmarks, as we will observe later, they exhibit inconsistent performance out-of-the-box on semantically rich ones.
The addition of CASE-enriched features establishes a new performance ceiling. 
By bridging the ``semantic gap'' for these models, our approach successfully combines the predictive power of tabular ICLs with semantic reasoning of LLMs.

\subsection{In-Depth Analysis of CASE}
Next, we study the overall performance of CASE on CARTE, TextTab, but also TabArena, and evaluate the synergy of CASE in combination with TabICLv2~\cite{qu2026tabiclv2} as the overall best-performing model from our previous evaluation. 
For preprocessing, we adhere to TabICL's recommended configurations of Skrub's \texttt{TableVectorizer} to ensure a fair comparison with established statistical pipelines. 
The main results are summarized in Table \ref{tab:tabular-benchmarks}.
We provide additional details and investigate the statistical significance of our findings in Appendix~\ref{app:add-results}.

\custompar{Performance on Semantic Benchmarks} 
As previously indicated, on the semantically rich CARTE and TextTab benchmarks, our approach largely outperforms all existing baselines. 
On CARTE, we observe a leading gap of over 2 pp in accuracy and almost 4 pp in R\textsuperscript{2} over AutoGluon, which ensembles and stacks a multitude of model architectures and involves significant training and tuning for each dataset. 
Similarly, on TextTab, TabICLv2 with CASE has a more than 2 pp lead in terms of accuracy while being competitive with SOTA on regression tasks.
This substantial margin of improvement over TabICLv2, AutoGluon, and ConTexTab suggests that, for datasets where column headers and textual features carry deep external meaning, the infusion of LLM-based world knowledge combined with a table-native target-aware training of the TLM is more effective than pure statistical feature engineering or isolated LLM embeddings.

\custompar{Boundaries of Semantic Priming} 
While showcasing significant improvements on semantically rich benchmarks, we observe a small performance trade-off on the TabArena benchmark, where the raw statistical learners (TabICLv2 and AutoGluon) maintain a lead. 
This is expected, as TabArena is a numerics-heavy suite where the primary predictive signals are numerical rather than semantic. 
While the drop in performance of TabICLv2 with CASE over the default TabICLv2 is small, it is statistically significant, as we investigate in more detail in Appendix~\ref{app:add-results}.
However, this also highlights the specialized nature of CASE: it acts as a modular semantic enhancer that can be used selectively. 
%In practice, our results suggest a clear heuristic: CASE provide maximum utility for tables with high-cardinality text or domain-specific entities, while standard statistical learners remain optimal for predominantly numerical data.

\custompar{Sample Efficiency and Few-Shot Performance} 
To investigate the effectiveness of CASE in the few-shot and low-data regime, we provide a comparison using subsampled training splits of the CARTE benchmark.
That is, we subsampled the CARTE training splits ranging from \num{128} to \num{8}k rows while keeping the test split untouched. To rigorously account for the inherent variance of random context selection, each evaluation is repeated over 5 independent runs using different random seeds. 

\begin{figure*}[t] %[hb]
\centering
\begin{subfigure}[t]{0.5\textwidth}
    \includegraphics[width=\linewidth]{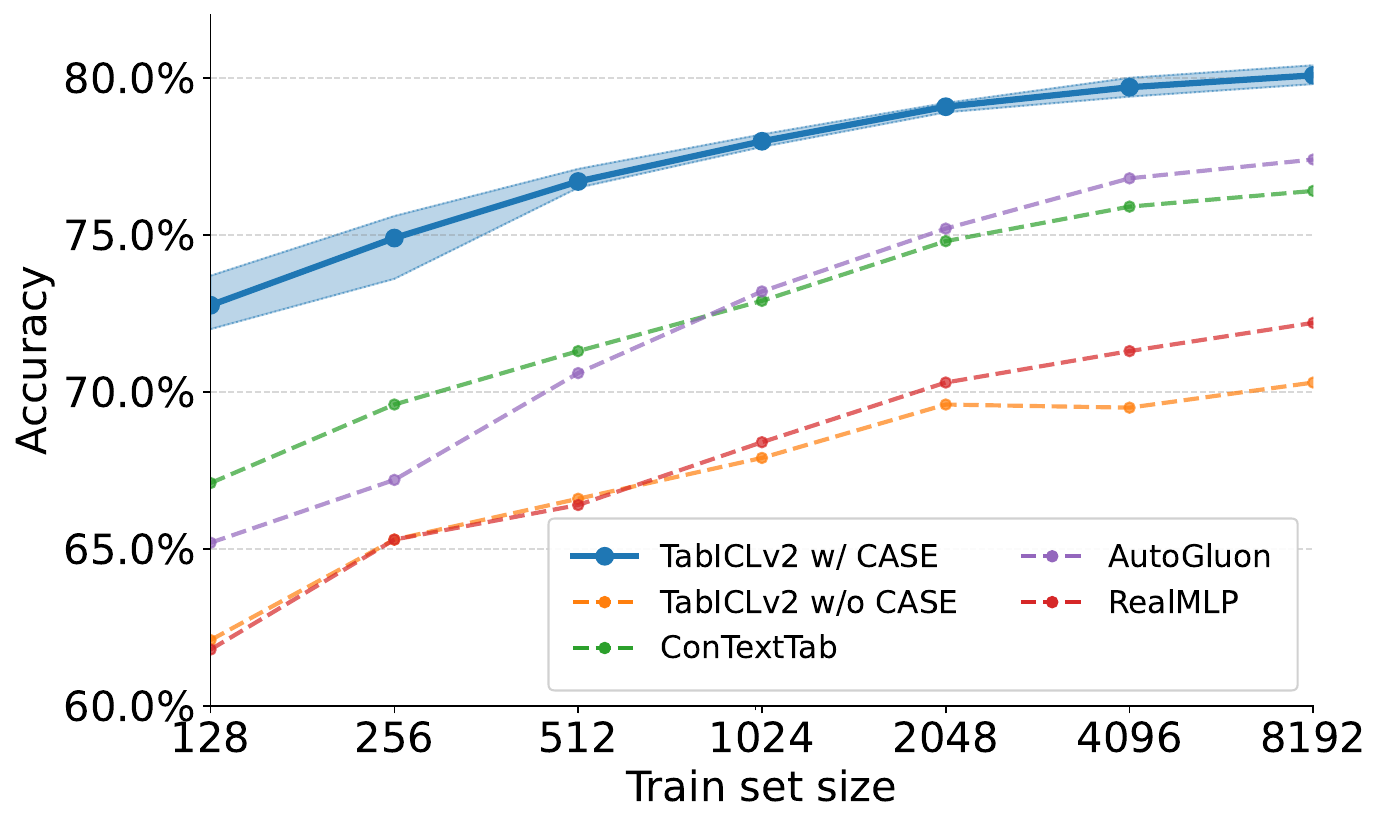} 
    \caption{Average accuracy across classification tasks.}
\end{subfigure}\hfill
\begin{subfigure}[t]{0.5\textwidth}
    \includegraphics[width=\linewidth]{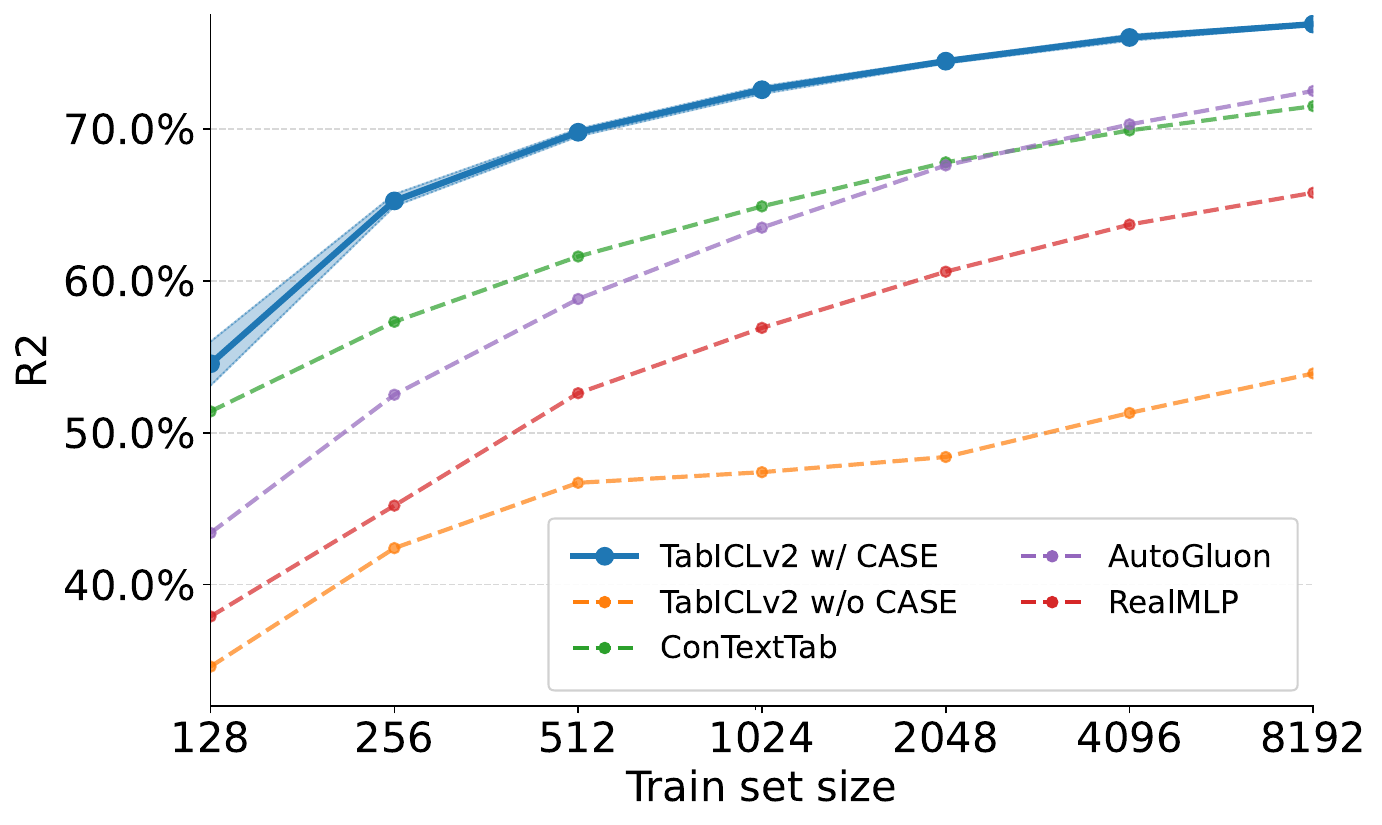}
    \caption{Average R$^\textrm{2}$ across regression tasks.}
\end{subfigure}
\caption{Impact of Context-Aware Semantic Embeddings (CASE) on predictive performance of varying subsamples of the train splits from the CARTE benchmark using a fixed test set. Results for CASE correspond to mean (solid) and min/max (shaded) over five different seeds for context priming.%
%While TabGemma struggles with regression, its latent semantic features act as a high-fidelity substrate for TabICL. This synergy allows TabICL plus CASE to achieve superior prediction quality with significantly fewer training samples than other tabular learners.
}
\label{fig:subset}
\end{figure*}

Our results, shown in Figure~\ref{fig:subset}, indicate that in low-data regimes, CASE is highly dominant, outperforming state-of-the-art baselines by a large margin. 
Furthermore, the performance loss for smaller train sizes seems to be less strong as for conventional predictors.
This suggests that the semantic priors embedded in the KV-primed representations provide a critical advantage when statistical signals are sparse, likely benefiting from the pretrained LLM's world knowledge. 
Furthermore, we observe a powerful synergy between the constituent models: the TLM generates rich, high-dimensional latent representations that can be effectively leveraged by the tabular learning model. 
%Thus, CASE successfully reconcile the semantic reasoning of LLMs with the robust statistical capabilities of TabICLv2, resulting in a highly synergetic hybrid approach.

\subsection{Ablations}
We conduct a series of ablation experiments on the CARTE and TextTab benchmarks to validate our core hypotheses regarding the model architecture, parameter and context scaling, as well as feature representations.
Throughout, we use TabICLv2 as the final predictor, as previously discussed.

\begin{figure*}[t] %[hb]
\centering
\begin{subfigure}[t]{0.48\textwidth}
    \includegraphics[width=\linewidth]{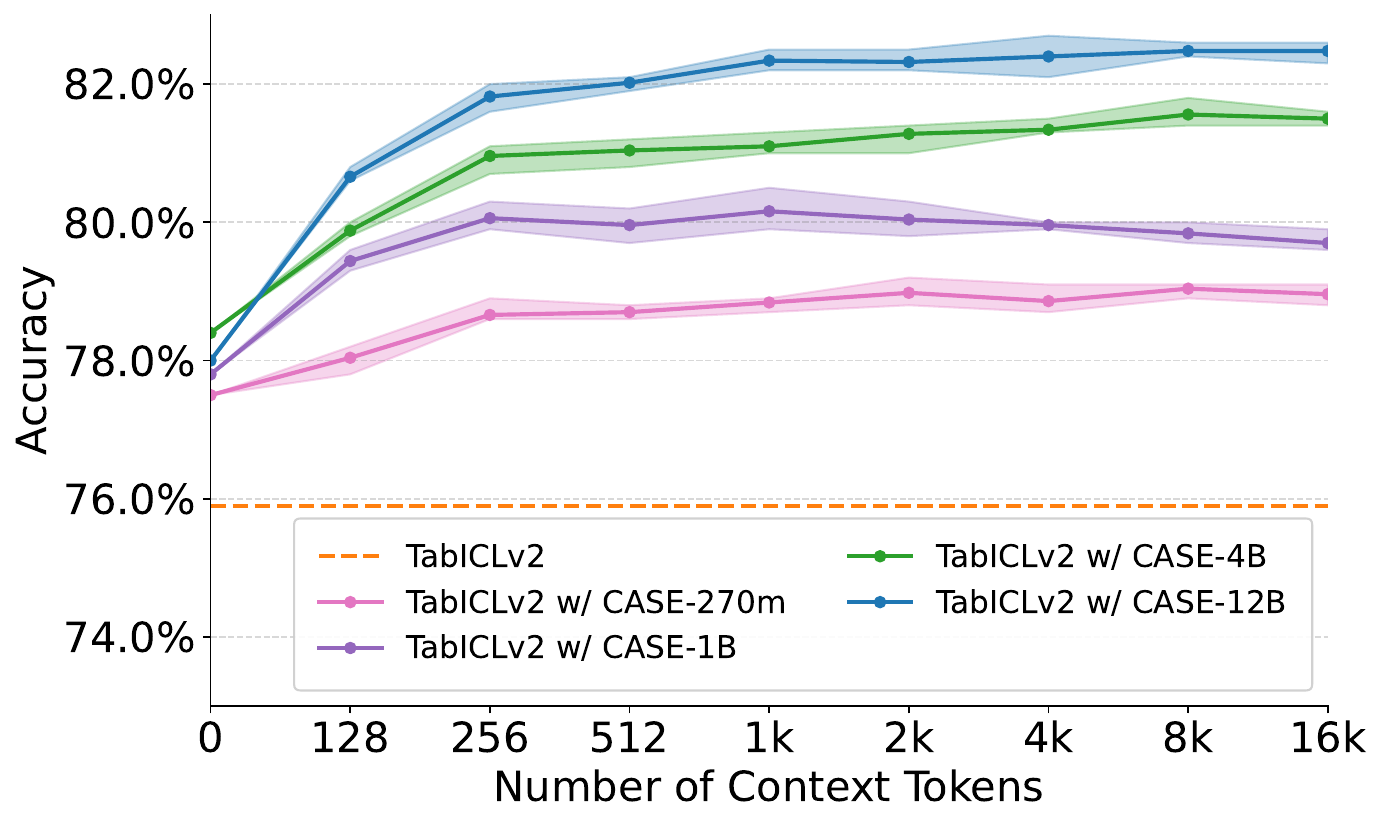} 
    \caption{Average accuracy across classification tasks.}
\end{subfigure}\hfill
\begin{subfigure}[t]{0.48\textwidth}
    \includegraphics[width=\linewidth]{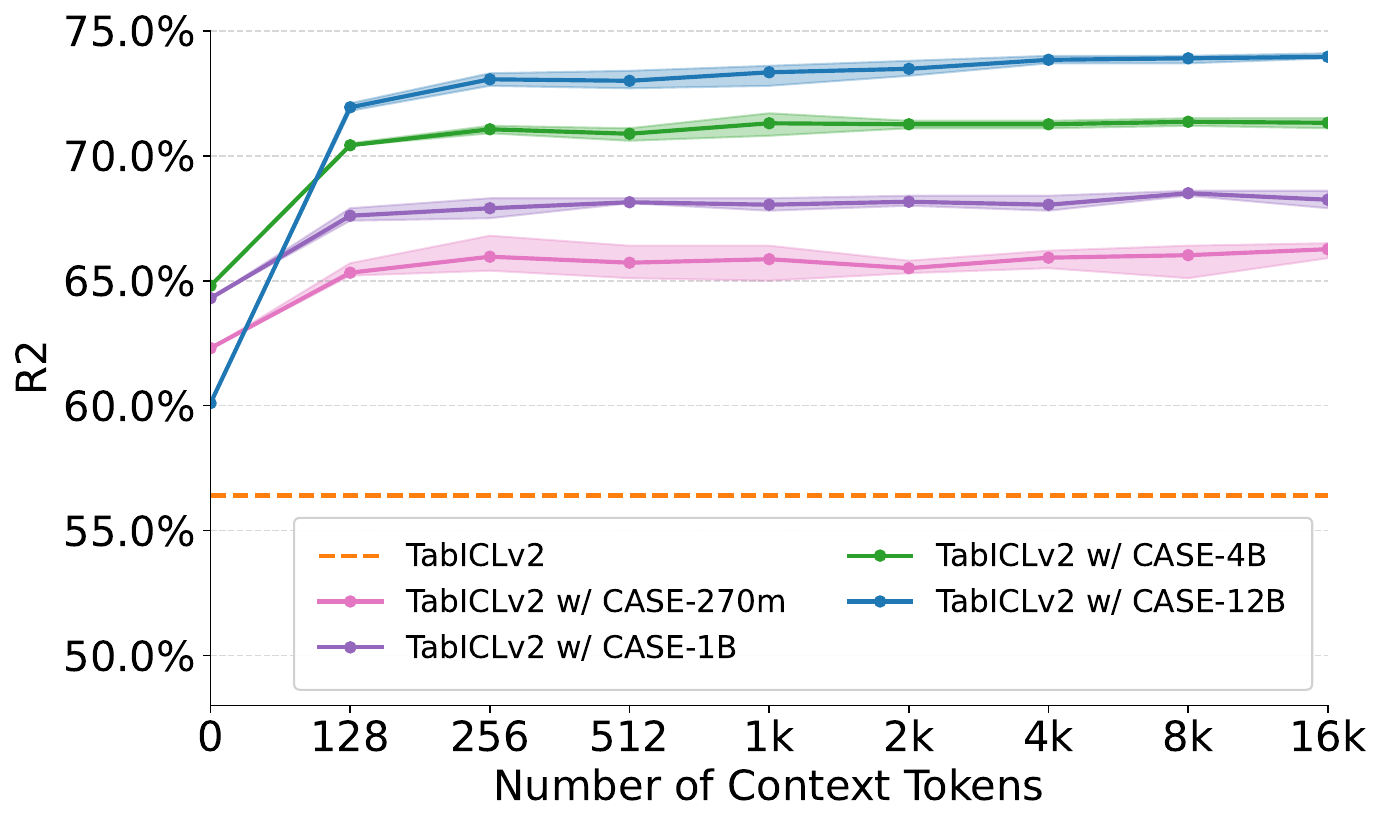}
    \caption{Average R\textsuperscript{2} across regression tasks.}
\end{subfigure}
\caption{Performance across CARTE and TextTab for TabICLv2 enriched with CASE embeddings across varying parameter counts and context window utilization.
Results correspond to mean (solid) and min/max (shaded) over five different seeds for context priming.}
\label{fig:model_scaling}
\end{figure*}

% \subsubsection{Impact of Scaling and Architecture}
To understand the scaling laws governing CASE, we investigate how both parameter count and context length influence the quality of the resulting embeddings. 
In Figure~\ref{fig:model_scaling}, we evaluate variants across four model scales -- 270M, 1B, 4B, and 12B parameters -- and nine context depths, ranging from 128 to 16k tokens (including a zero-context baseline). Again, we account for the variance of random context selection by evaluating over 5 independent runs using different random seeds.

\custompar{Model Scaling}
Our results reveal that all CASE configurations -- regardless of base model scale or utilized context -- consistently outperform the vanilla TabICLv2 baseline. %, even in the uncontextualized case of 0 context tokens. 
We observe a robust, monotonic relationship between parameter count of the TLM base model and predictive performance: as the model size increases from 270M to 12B, the quality of the semantic embeddings improves across both classification and regression tasks as reflected by the increased performance. 
This aligns with established scaling laws for LLMs, suggesting that larger models possess a richer latent ``world model'' that allows for more precise mapping of tabular values to their underlying concepts.

\custompar{Context Scaling}
Across all base TLM variants, we observe a significant leap in performance when transitioning from 0 context tokens (corresponding to uncontextualized row embeddings) to 1k tokens. 
This ``contextual leap'' validates our hypothesis that a minimal semantic anchor is necessary to resolve the inherent ambiguity of isolated rows or cell values. 
Interestingly, even uncontextualized embeddings (0 context tokens) provide a substantial boost over the TabICLv2 baseline, indicating that the zero-shot semantic recognition of individual features is, on its own, already valuable.

\custompar{Robustness to Random Selection}
Crucially, each of our 5 independent evaluation runs draws a completely distinct subset of context rows with entirely randomized permutations in row ordering. Despite this shuffling, the resulting min/max variance bands visualized in Figure~\ref{fig:model_scaling} are notably tight. The marginal fluctuations in downstream performance demonstrate that CASE is remarkably stable and resilient to the stochasticity of context row selection. This indicates that the model successfully distills the global data distributions and semantic structure from the anchor data regardless of the specific layout or samples encountered during inference.

\custompar{Saturation and Complementary Reasoning}
While performance continues to slightly improve as context scales up to 32k tokens, we observe a diminishing marginal return beyond the initial 1k boost. 
This is somewhat unexpected. As larger models should be able to leverage much larger context windows, one would have expected this saturation point to shift towards larger number of context tokens for larger base model sizes.
Maybe, this saturation suggests a synergy between the embedding model and the tabular learner: 
Our tuned TLM acts as a high-level ``reasoner'' that requires only a small representative sample of rows to ground the tables' semantics. 
Once the model has resolved the individual row's context, the statistical heavy lifting is effectively offloaded to the downstream tabular learner, which processes the full training set. 
This synergy allows CASE to remain computationally efficient, as it only requires a comparably small context KV cache.
However, these findings may also suggest further room for improvements, e.g.\ in pretraining, to more effectively leverage the larger context windows of the larger TLM base models.

\begin{figure}
\begin{minipage}[t]{0.57\textwidth}
  \vspace{0pt}
  \centering
  \includegraphics[trim=0 0 0 7mm, clip, width=\linewidth]{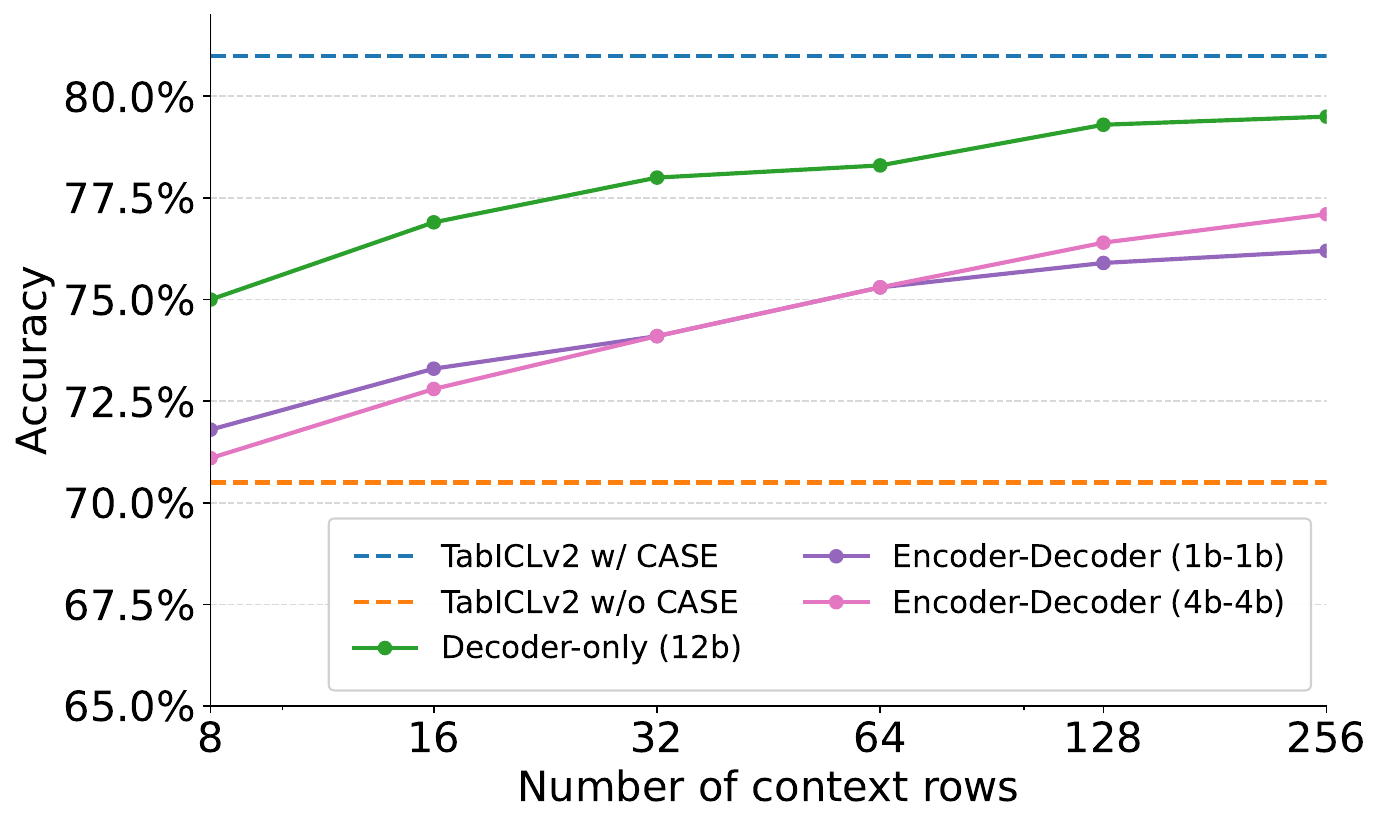}
  \captionof{figure}{Architectural Comparison on CARTE classification tasks, comparing the predictive capabilities of table-tuned LLMs with different architectures.}
  \label{fig:dec_vs_enc}
\end{minipage}
\hfill
\begin{minipage}[t]{0.4\textwidth}
  \vspace{0pt}
  \centering
  \footnotesize
  \setlength{\tabcolsep}{2.5pt}
  \captionof{table}{Feature encoding ablations results on the CARTE and TextTab benchmarks, depicting mean accuracy for classification and $R^2$ for regression tasks, as well as mean rank across all tasks.}
  \label{tab:tabular-ablations}
  % \begin{table}[t]\centering\footnotesize
% \caption{Ablations on the CARTE and TextTab benchmarks.}
% \label{tab:tabular-ablations}
\begin{tabular}{lrrr}
\toprule
% & \multicolumn{3}{c}{\textbf{Ablation}} & \\ \cmidrule(lr){2-4}
Model & \multicolumn{1}{c}{\textbf{Rank}} & \multicolumn{1}{c}{\textbf{Acc}} & \multicolumn{1}{c}{\textbf{R$^\textrm{2}$}}  \\
\midrule
\bfseries TabICL [TF-IDF + CASE] & \bfseries 1.97 & \bfseries 83.1 & \bfseries 75.5  \\
 TabICL [Default + CASE]  & 3.38 & 82.5 & 73.9 \\
 TabICL [Row-Embedding] & 3.52 & 81.7 & 73.7 \\
 TabICL [Cell-Embedding] & 4.28 &  81.2 &  72.0  \\
 TabICL [CASE w/o $X_{\textrm{org}}$] & 4.41 & 82.1 & 71.7 \\
 TabICL [Gemma3] & 4.48 &  81.8 & 72.5  \\
 TabICL [TF-IDF] & 4.76 &  80.9 & 69.7  \\
 TabICL [Default] & 7.06 & 75.9 & 56.4  \\
%\\
%Gemma 3 12B & \NAN & \NAN & 4.6 & -98.4 & \NAN & \NAN & \NAN & \NAN \\
\bottomrule
\end{tabular}
% \end{table} %\v
\end{minipage}
\end{figure}

\custompar{Standalone Generative Prediction and Architectural Analysis}
\label{sec:dec_vs_enc}
While CASE is primarily designed to augment downstream tabular learners, the generative pre-training of the underlying TLM makes the model also usable as a standalone few-shot predictor. 
In Figure~\ref{fig:dec_vs_enc}, we evaluate this capability by comparing the vanilla TabICLv2, the CASE-enriched TabICLv2, and the standalone CASE-TLM (decoder-only 12B) on CARTE classification tasks. 

As a standalone model, the decoder-only TLM demonstrates strong predictive performance that scales monotonically with the number of context rows. 
Notably, given only 8 context rows, our TLM outperforms TabICLv2 (given the full train split) across the CARTE classification tasks.
However, note that CARTE contains exclusively binary classification targets.
This confirms that the model has successfully learned to predict target tokens from its context. 
However, the TLM stays behind TabICLv2 with CASE, confirming the complementary nature of the two approaches.
%The hybrid CASE approach combines the high-level semantic reasoning of the LLM with the statistical capabilities of the tabular learner, effectively bypassing the context limitations of generative models. 
Note that standalone regression was not evaluated, as standard decoder-only models do not natively support continuous value regression without specialized head fine-tuning.

Moreover, we compare different LLM architectures: In the literature, encoder-decoder (e.g., T5) or encoder-only (e.g., BERT) architectures are more conventional choices for embedding generation.
To validate our choice of a decoder-only backbone, we trained encoder-decoder variants using the \textit{T5Gemma2} architecture~\cite{zhang2025t5gemma2seeingreading} under an identical training budget and training objective.
As illustrated in Figure~\ref{fig:dec_vs_enc}, while the encoder-decoder variants also scale with context, they consistently and largely under perform relative to the decoder-only models.
%This may be due to architectural limitations or the generally better pretrained checkpoints of the used Gemma 3 model.

% \input{tables/ablation_table}

% \begin{figure*}
%     \centering
%     \includegraphics[scale=0.4]{figs/enc_dec_carte.pdf}
%     \caption{\textbf{Architectural Comparison on CARTE Classification.} We compare the standalone predictive performance of decoder-only vs. encoder-decoder architectures across varying context row counts, alongside the hybrid CASE+TabICLv2 approach.}
%     \label{fig:dec_vs_enc}
% \end{figure*}

\custompar{Impact of Contextualization} To isolate the benefit of the proposed KV-cache priming, we compare CASE ($k\,{=}\,128$) against a non-contextualized row-embedding baseline (\texttt{TabICL[Row-Embedding]}, $k\,{=}\,0$) as well as cell-level embeddings (\texttt{TabICL [Cell-Embedding]}) obtained via a standard Sentence Transformer, \texttt{all-MiniLM-L6-v2}, similar to related works~\cite{contexttab}.
We compare this against the default TabICLv2 model as well as combining TabICLv2 with Skrub's TF-IDF-based featurizer.
The results are depicted in Table~\ref{tab:tabular-ablations}.

The significantly degraded performance of the non-contextualized variant as compared to the full CASE approach confirms that row-level semantics are inherently distributional. 
Without the provided semantic context, the model fails to resolve the semantic ambiguities potentially present in an isolated row.
Similarly, using cell-level embeddings shows marginal improvements over TF-IDF-based embeddings, but consistently lags behind our contextualized row embedding approach. 
%This suggests that the inter-feature relationships captured via CASE are significantly more informative than aggregated independent cell vectors.

\custompar{Importance of Tabular-Specific LLMs}
Further, we replaced the specialized table-tuned TLM backbone with an off-the-shelf Gemma-3-12B model \cite{gemma3}, validating our pretraining objective and table-native tuning. 
The results are again depicted in Table~\ref{tab:tabular-ablations}.
While again showing slight improvements as compared to conventional TF-IDF features, the significant performance drop compared to CASE underscores that generic LLMs, while possessing vast world knowledge, require table-native adaptations to correctly interpret the structural and relational nuances of tabular data.

\custompar{Role of Statistical Preprocessing} Finally, we evaluate the synergy between traditional preprocessing and semantic enhancement by comparing TabICLv2 with and without TF-IDF features using Skrub's \texttt{TableVectorizer}.
Our results demonstrate that while classical statistical techniques remain a robust baseline, the addition of CASE consistently yields performance gains regardless of the preprocessing used. 
To isolate the contribution of our embeddings, we further evaluate a variant using only CASE features (\texttt{TabICL [CASE w/o $X_{\textrm{org}}$]}). We find that standalone embeddings serve as high-utility features; however, the inclusion of the original $X_{\text{org}}$ features yields the strongest results. %This result highlights the complementary nature of CASE's semantic world knowledge and the raw features' direct statistical signal.

\section{Conclusions}
\custompar{Limitations and Future Work}
While CASE demonstrate significant potential, our current implementation has some limitations to be addressed in future works.
First, as our framework is built upon a table-tuned Gemma model, it inherits constraints applicable to all table-serialization based approaches combined with LLMs: a lack of permutation equivariance with respect to both row and column ordering (which we, however, find empirically not to be statistically significant in Appendix~\ref{sec:appendix-column-permutation}), and the limited numerical inductive bias inherent in token-based language models. Also, our current context priming strategy relies on uniform random sampling to populate the KV cache, which is inherently stochastic and may omit critical anchor rows that define the table's distribution. However, we empirically find the random context sampling not to be statistically significant (see~Appendix~\ref{sec:appendix-context-permutation}). %The performance of CASE could likely be further improved by employing more sophisticated row-selection heuristics.
Nevertheless, this motivates further research into developing LLM architectures specifically optimized for tabular structures, for example adapting recent approaches for permutation-invariant attention patterns and positional encodings~\cite{egressy2025set}. Furthermore, using PCA to project the high-dimensional embeddings into a compact latent space may be suboptimal. Future works could investigate non-linear manifold learning techniques which may better preserve the subtle semantic variances in the data.
While the use of a language model increases the computational footprint during feature extraction, our KV-cache priming strategy ensures that this overhead remains significantly lower than traditional AutoML pipelines or exhaustive hyperparameter searches (see Appendix~\ref{app:efficiency}).
Finally, while utilizing open-source benchmarks carries a theoretical risk of data contamination from the LLM backbone or continued pretraining, our empirical analyses suggest that performance gains are driven by genuine semantic reasoning rather than memorization (see Appendix~\ref{sec:appendix-semantics}). We discuss potential data contamination in detail in Appendix~\ref{app:contamination} and further validate our findings in Appendix~\ref{sec:appendix_strable} on the recently introduced STRABLE benchmark~\cite{strable}, which corroborates the strong performance improvements observed on CARTE and TextTab.

\custompar{Summary}
In this work, we introduced CASE, a framework designed to bridge the historical divide between the vast semantic world knowledge of LLMs and the rigorous statistical inductive biases of specialized tabular learners. At the heart of our contribution is a pretrained, table-tuned LLM which functions as a few-shot semantic engine for structured data. By employing a novel KV-cache priming strategy, CASE enables the generation of target-aware row representations that are no longer processed in a semantic vacuum, but are instead anchored within the specific statistical and domain distribution of the parent table.
Our empirical evaluation across the CARTE and TextTab benchmarks demonstrates that CASE substantially advances performance baselines for semantically rich tabular data. The framework is particularly potent in the low-data regime, where the LLM’s learned priors compensate for the lack of sufficient statistical signals.
By demonstrating that decoder-only architectures can serve as efficient, context-aware embedding engines, we provide a scalable blueprint for building foundation models for tabular data.

\section*{Acknowledgements}
We would like to thank Johannes Hoffart, Markus Kohler, and Sam Thelin for their insightful comments and suggestions throughout the development of this work. We thank Myung Jun Kim for providing valuable feedback on experimental setups and drafts of this contribution.
We thank Michael Medek for implementation support in initial versions of our Tabular Language Model.

{
\small
\bibliography{ref}
\bibliographystyle{plainnat}
}

\clearpage
\newpage

%%%%%%%%%%%%%%%%%%%%%%%%%%%%%%%%%%%%%%%%%%%%%%%%%%%%%%%%%%%%

\appendix

\section{Baseline Details}\label{app:baselines}
\custompar{TabICLv2}
We use the model from the official \texttt{tabicl} package at version 2.0.3 using its default feature preprocessing if not marked otherwise specifically.

\custompar{TabPFN}
We use the model from the official \texttt{tabpfn} package at version 7.1.1, corresponding to the TabPFN-2.6 checkpoint, using its default feature preprocessing if not marked otherwise specifically.

\custompar{AutoGluon}
We evaluate AutoGluon v1.5 with its native feature encoder. We use the \texttt{best\_quality} preset with a per-dataset time limit of 4\,h running on a 32 cores machine with 256\,GB of RAM.

\custompar{ConTextTab}
We evaluate ConTextTab v1.1.2 using the reference implementation and checkpoint\footnote{\href{https://github.com/SAP-samples/contexttab}{github.com/SAP-samples/contexttab}}. We set a context size of 8k samples and evaluate with 8-fold bagging.

\custompar{Pytabkit models} 
We use the \texttt{pytabkit}~\cite{realmlp} for evaluating XGBoost, CatBoost, and RealMLP. 
We evaluate these models with ensembled hyperparameter optimization across 5-fold inner cross-validation (HPO-CV).
For the HPO variants, we use the recently added \texttt{tabarena} search spaces proposed in~\cite{tabarena}.

\custompar{Naive}
We use the Naive predictor from sklearn v1.5.2, estimating the median for regression tasks or the majority class in the case of classification tasks.

\subsection{Feature preprocessing}\label{app:preprocessing}
If available, we use the default feature preprocessing of the individual baselines investigated. 
For Naive, XGBoost, CatBoost, and RealMLP, we use the preprocessor from AutoGluon.
For TF-IDF features, we use Skrub's \texttt{TableVectorizer}, which uses a pass-through for low-cardinal and numerical features to be handled natively by TabICL, and the TF-IDF-based \texttt{StringEncoder} for high-cardinal features with more than 40 classes, using a fixed seed.
Note that the TableVectorizer also applies PCA to the obtained features, downcasting them to 30 dimensions.
Finally, note that these features are calculate per column. That is, if the table contains 10 high-cardinal or free text features, the approach appends 300 additional features after the transform.

\section{Additional Results}\label{app:add-results}
In the following, we provide additional results on both the semantically rich benchmarks (CARTE and TextTab) as well as the numerics-heavy TabArena.
We report critical difference (CD) diagrams, as well as win-ratios and ELO scores.

For the calculation of the Elo scores, we utilize the implementation of TabArena~\cite{tabarena}, which itself is based on ChatBot Arena's implementation~\cite{chiang2024chatbot}.

For all ranking results (aside from those used in the critical difference diagrams), we use a robust ranking algorithm.
That is, models that lie within \num{0.1} percentage points (in terms of either accuracy or $R^2$) are considered ties.
The same robustification also applies to our Elo score and win-ratio calculations.

For the critical difference (CD) diagrams, we use the \texttt{autorank} library~\cite{autorank}.
So as not to influence the statistical guarantees of the significance test used, we do not alter or robustify its internal rank calculation.
Hence, the ranks shown in the CD diagrams can slightly deviate from those shown in the result tables or figures.

\newpage
\subsection{Semantically Rich Benchmarks}

\begin{figure}
\centering
\begin{subfigure}[t]{0.9\textwidth}\centering
    \includegraphics[trim={20mm 0 0 0}, clip, width=\linewidth]{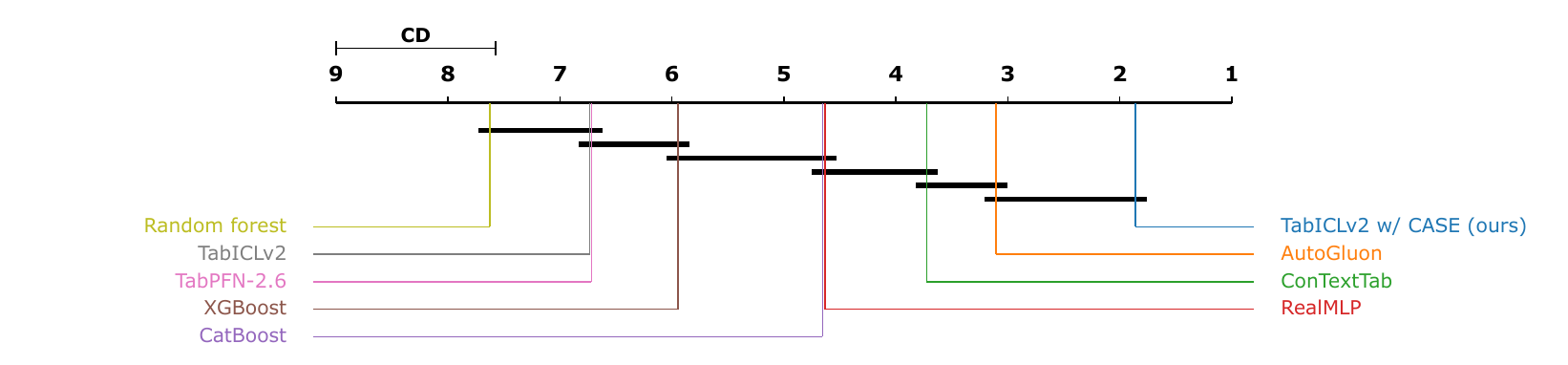}
    \caption{Critical difference diagram.}
\end{subfigure}\\[4mm]
\begin{subfigure}[t]{0.45\textwidth}\centering
    \includegraphics[trim=0 15mm 0 0, clip, width=\linewidth]{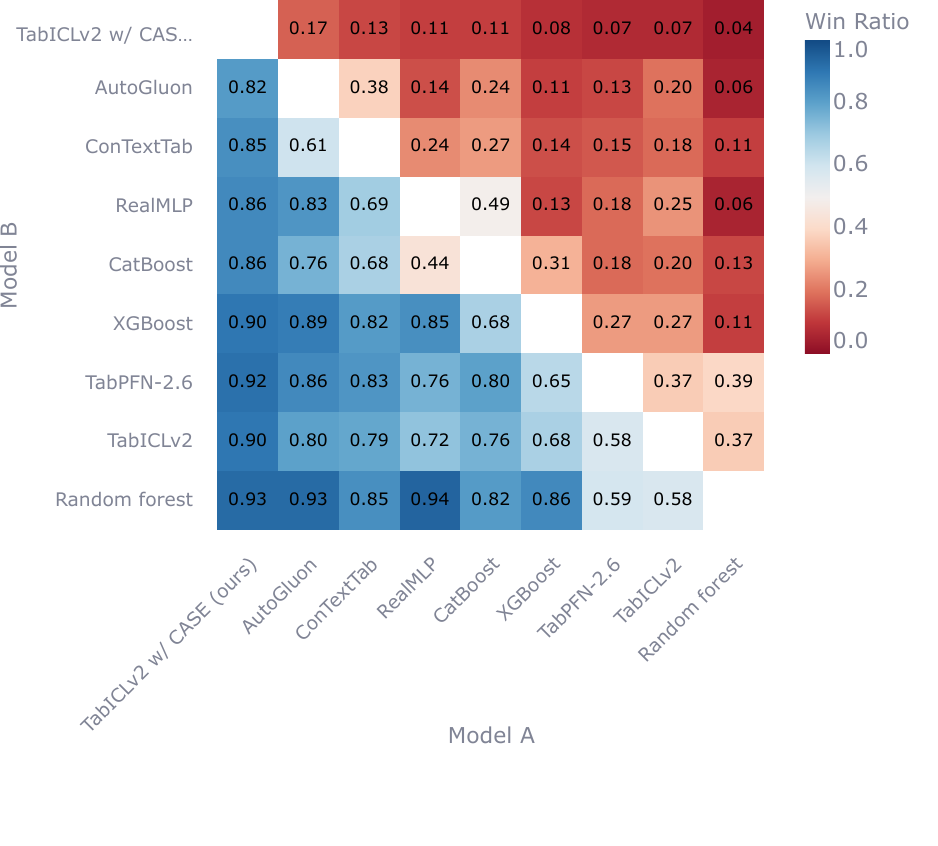} 
    \caption{Win ratios.}
\end{subfigure}
\hfill
\begin{subfigure}[t]{0.54\textwidth}\centering
    \includegraphics[trim=0 15mm 0 0, clip, width=\linewidth]{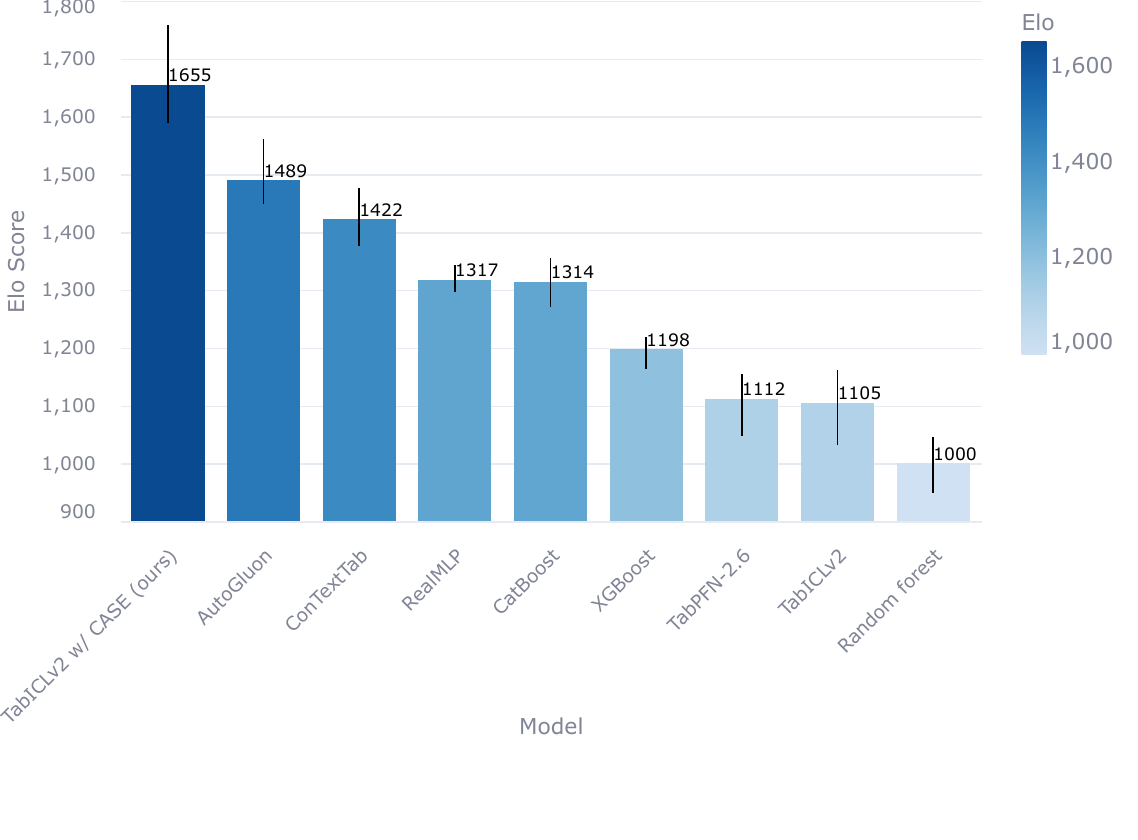} 
    \caption{ELO scores.}
\end{subfigure}
\caption{Critical difference diagrams, win ratios and ELO scores for the semantic-heavy CARTE and TextTab benchmarks. ELO scores are normalized to Random Forest at 1000 ELO.}
\label{fig:extended-results-semantic-rich}
\end{figure}

\begin{figure}
\centering
\includegraphics[width=\linewidth]{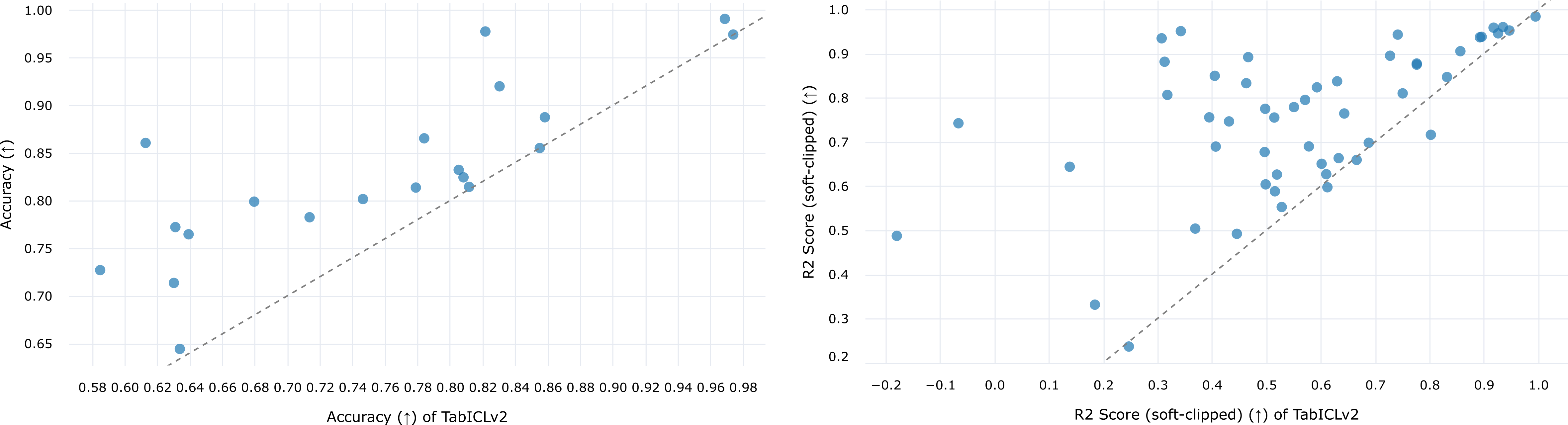} 
\caption{Per-task accuracy (left) and R\textsuperscript{2} scores (right) on CARTE and TextTab comparing TabICLv2 with (y-axis) and without CASE (x-axis). On-diagonal scores indicate identical performance while above-diagonal scores indicate that performance is better when using CASE instead of the default.}
\label{fig:extended-results-semantic-rich-1v1}
\end{figure}

For the CARTE and TextTab benchmarks, the CD diagram, win ratios and ELO score plots are depicted in Figure~\ref{fig:extended-results-semantic-rich}.
As before, we observe that TabICL with our proposed CASE enrichment significantly outperforms all baselines.
For all single-model baselines, this difference is statistically significant.
While our approach also outperforms AutoGluon, as previously denoted in terms of absolute performance and mean rank, now also shown in terms of win ratio and Elo scores, the difference is not quite statistically significant, as denoted in the CD diagram.
However, note that AutoGluon ensembles and stacks a multitude of models, requiring extensive tuning for each dataset, while TabICL with CASE performs on-the-fly predictions following the in-context paradigm.
Finally, a one-vs-one comparison of per-task results using TabICLv2 with and without CASE, shown in Figure~\ref{fig:extended-results-semantic-rich-1v1}, show that the gains in absolute performance are substantial in many cases.

Overall, also these extended results underline the superior performance of our approach, setting a new state of the art for semantically rich datasets.

\subsection{TabArena}

\begin{figure}[t]
\centering
\begin{subfigure}[t]{0.9\textwidth}\centering
    \includegraphics[trim={5mm 0 0 0}, clip, width=\linewidth]{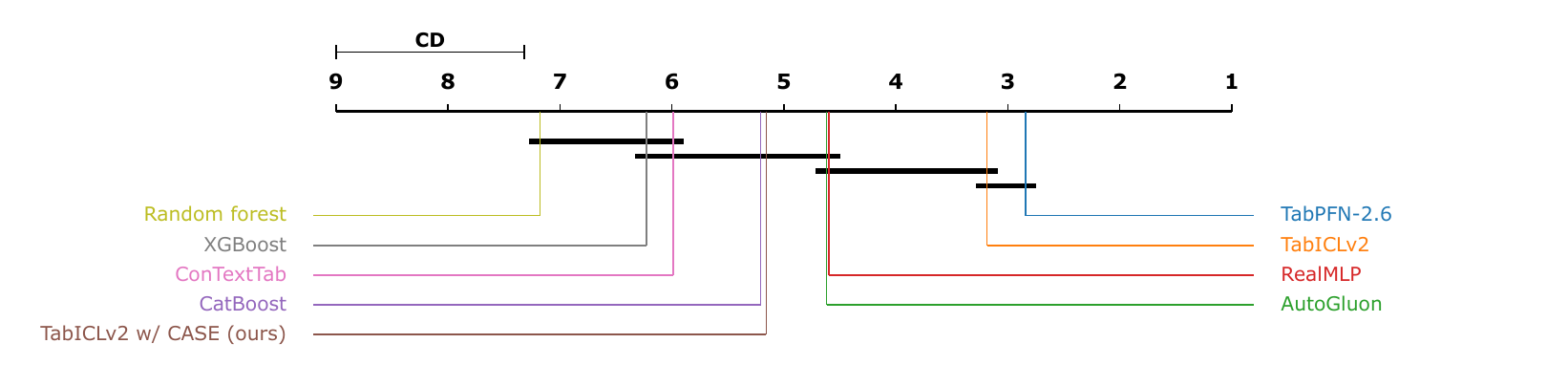}
    \caption{Critical difference diagram.}
\end{subfigure}\\[4mm]
\begin{subfigure}[t]{0.45\textwidth}\centering
    \includegraphics[trim=0 15mm 0 0, clip, width=\linewidth]{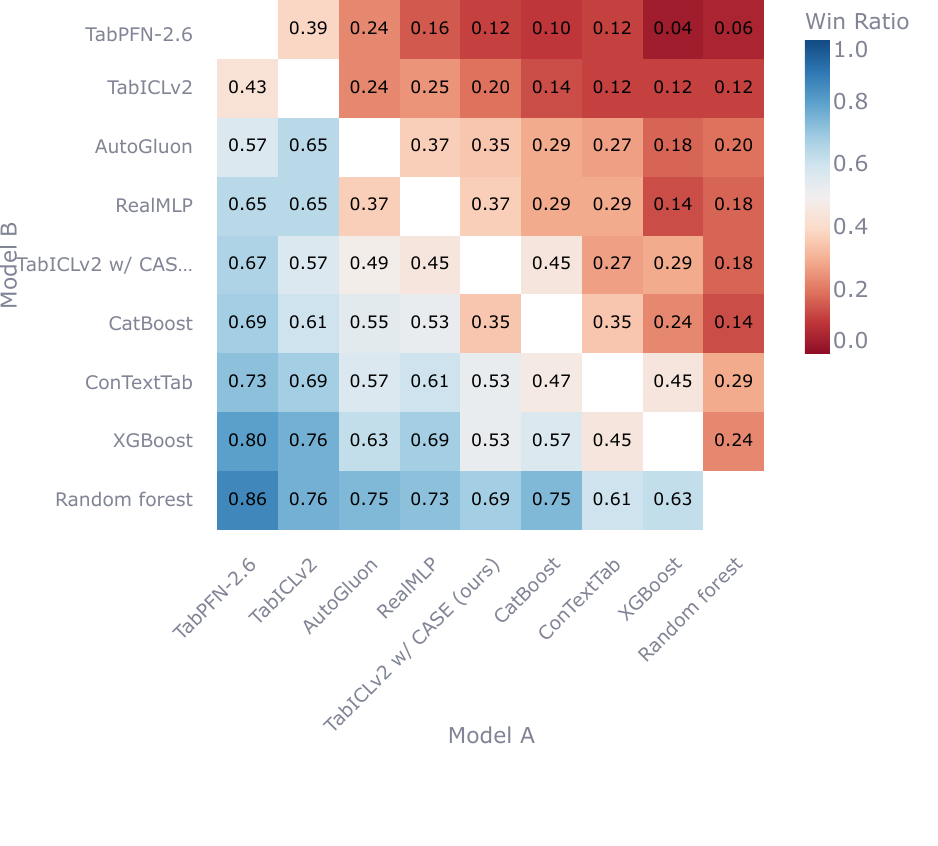} 
    \caption{Win ratios.}
\end{subfigure}
\hfill
\begin{subfigure}[t]{0.54\textwidth}\centering
    \includegraphics[trim=0 15mm 0 0, clip, width=1.0\linewidth]{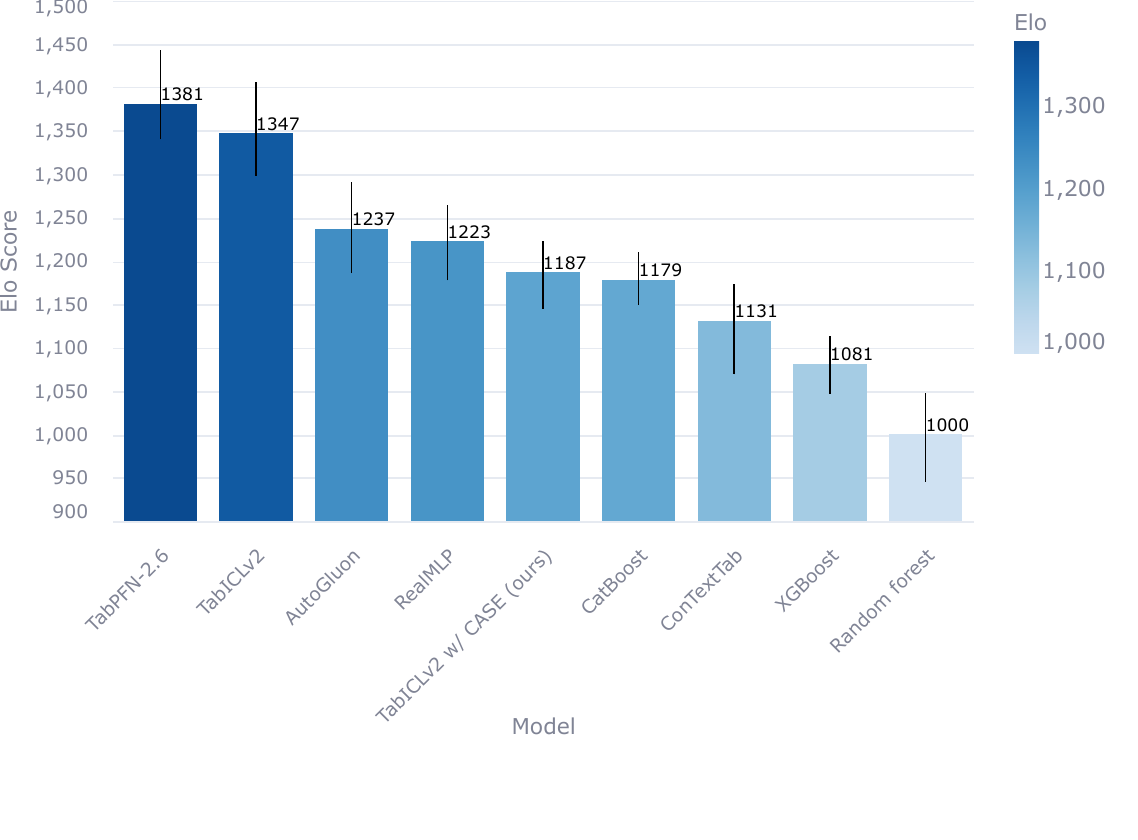} 
    \caption{ELO scores.}
\end{subfigure}
\caption{Critical difference diagrams, win ratios and ELO scores for the numerics-heavy TabArena benchmarks. ELO scores are normalized to Random Forest at 1000 ELO.}
\label{fig:extended-results-tabarena}
\end{figure}

\begin{figure}[t]
\centering
\includegraphics[width=\linewidth]{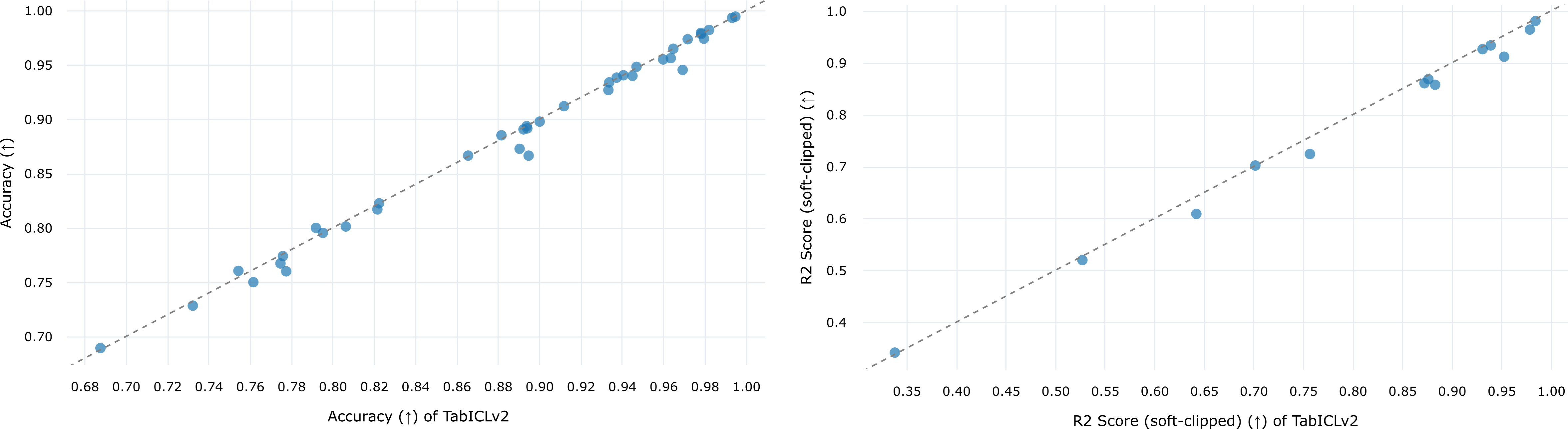} 
\caption{Per-task accuracy (left) and R\textsuperscript{2} scores (right) on TabArena comparing TabICLv2 with (y-axis) and without CASE (x-axis). On-diagonal scores indicate identical performance while above-diagonal scores indicate that performance is better when using CASE instead of the default.}
\label{fig:extended-results-tabarena-1v1}
\end{figure}

As a robustness check, we also show extended results for the numerics-heavy TabArena benchmark in Figure~\ref{fig:extended-results-tabarena}.
As previously discussed, we observe a performance drop when using CASE with numerics-heavy semantic-scarce dataset.
This drop is statistically significant, as shown in the CD diagram.
While we observe this drop throughout, in terms of mean rank as well as win ratios and Elo scores, these ranking metrics to some degree hide the differences in the absolute values of the performance metrics:
while we have seen substantial gains on semantically rich datasets also in terms of absolute scores, the absolute \emph{drop} in performance on TabArena is relatively small (albeit statistically significant), as shown in Figure~\ref{fig:extended-results-tabarena-1v1}.
This hints at the problem of reporting or relying on single metrics for model comparisons, in particular ranking- or win-based ones: a tiny, but consistent, improvement may result in much better rank or Elo scores while hiding that the absolute improvement may be incremental.

\newpage
\custompar{Additional investigation of performance drop}
Nevertheless, to investigate the potential mechanism underlying this drop in performance, we compare our approach with the performance of TabICL under addition of uninformative features.
Doing so, we can investigate to what extent the drop in performance is due to a noisy or uninformative nature of CASE embeddings under purely-numeric datasets.
To this end, we append 32 randomly sampled features to the context and queries, corresponding to the same amount of additional features when using CASE, and evaluate over 5 different random seeds.
The results are depicted in Figure~\ref{fig:extended-results-tabarena-cd-diagram-vs-random}.
Indeed, we can observe that addition of random features does degrade TabICL's performance, albeit this degradation is not statistically significant.
While overall TabICL with CASE is statistically significantly worse than bare TabICL across TabArena, as previously discussed, this does not stretch across all noisy feature ablations.
Looking more closely, we can make out that the significant part of the performance degradation is driven by the regression tasks on TabArena, while performance remains statistically indistinguishable across the classification parts, where also absolute performance remains virtually identical, dropping from 88.7\% to 88.3\% accuracy, whereas on regression we see a drop from 79.9\% to 78.5\% R$^2$.

To summarize, uninformative, noisy feature do result in some degradation of TabICL's performance and explain part of the drop in performance when using TabICL with CASE on TabArena, in particular on regression tasks.
However, some additional degradation can be observed with CASE embeddings on numerics-heavy datasets, in particular on regression tasks, that go beyond noisy embeddings. 
This warrants further research to robustify CASE embeddings on regression tasks and under the presence of dominantly numerical features.
While these drops are statistically significant, they are small in terms of absolute scores.

\begin{figure}[t]
\begin{subfigure}[t]{0.9\textwidth}\centering
\centering
\includegraphics[trim={0 0 5mm 0}, clip, width=\linewidth]{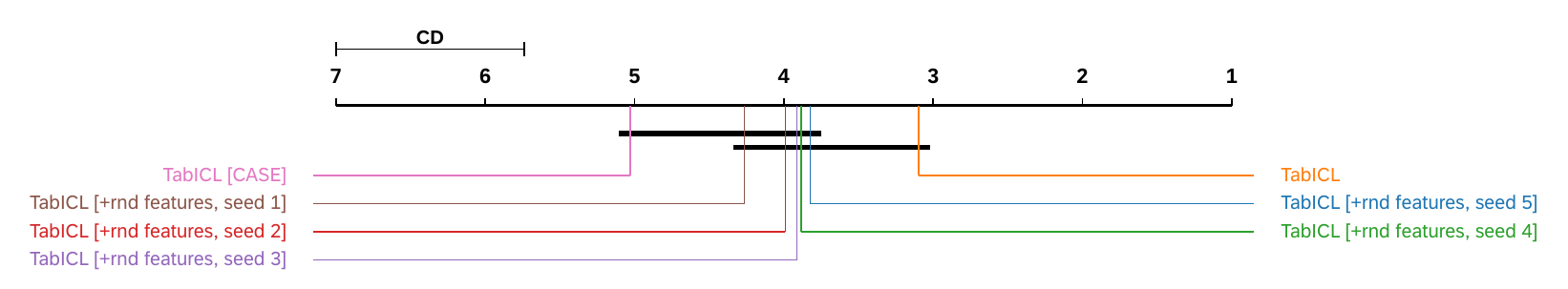} 
\caption{Across all tasks.}
\label{fig:extended-results-tabarena-cd-diagram-vs-random}
\end{subfigure}\\
\begin{subfigure}[t]{0.9\textwidth}\centering
\centering
\includegraphics[trim={0 18mm 5mm 0}, clip, width=\linewidth]{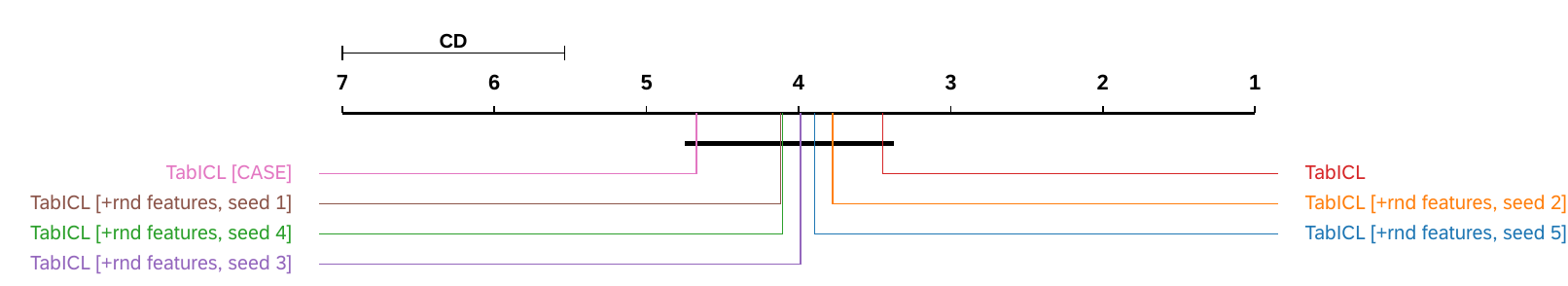} 
\caption{Across classification tasks only.}
\label{fig:extended-results-tabarena-cd-diagram-vs-random}
\end{subfigure}
\caption{Critical difference diagram comparing vanilla TabICL, TabICL with CASE and TabICL with additional random features across the TabArena benchmark.}
\end{figure}

%%%%%%%%%%%%%%%%%%%
%%%%%%%%%%%%%%%%%%%
%%%% ABLATIONS
%%%%%%%%%%%%%%%%%%%
%%%%%%%%%%%%%%%%%%%
\newpage \clearpage
\section{Additional Ablations}

%%%%%%%%%%
%%%% SEMANTIC CORRPUTION
%%%%%%%%%%

\begin{table}[t]
\centering
\footnotesize
\caption{Semantic corruption performance degradation: Disruption of textual semantics drops TabICL [CASE] back to pure non-semantic baseline performance (TabICL), confirming that CASE gains its predictive power via genuine semantic reasoning.}
\label{tab:semantic-corruption-ablation}
\begin{tabular}{l S[table-format=1.1] S[table-format=1.1] S[table-format=2.1] S[table-format=2.1] S[table-format=1.1] S[table-format=2.1] S[table-format=2.1]}
\toprule
& \multicolumn{1}{c}{All} & \multicolumn{3}{c}{CARTE} & \multicolumn{3}{c}{TextTab} \\
\cmidrule(lr){2-2} \cmidrule(lr){3-5} \cmidrule(lr){6-8}
Model & {Rank} & {Rank} & {Acc (\%)} & {$R^2$ (\%)} & {Rank} & {Acc (\%)} & {$R^2$ (\%)} \\
\midrule
TabICL [CASE]              & \bfseries 1.3 & \bfseries 1.3 & \bfseries 80.8 & \bfseries 77.5 & \bfseries 1.6 & \bfseries 86.2 & \bfseries 68.9 \\
TabICL                     & 2.3           & 2.2           & 78.3           & 73.9           & 2.5           & 84.1           & 54.5           \\
TabICL [corrupted]         & 2.7           & 2.8           & 77.7           & 73.4           & 2.4           & 84.0           & 63.8           \\
TabICL [CASE, corrupted]   & 3.0           & 3.2           & 78.2           & 72.2           & 2.5           & 84.2           & 63.6           \\
\bottomrule
\end{tabular}
\end{table}

\begin{figure*}[t]
    \centering
    \includegraphics[width=0.9\linewidth]{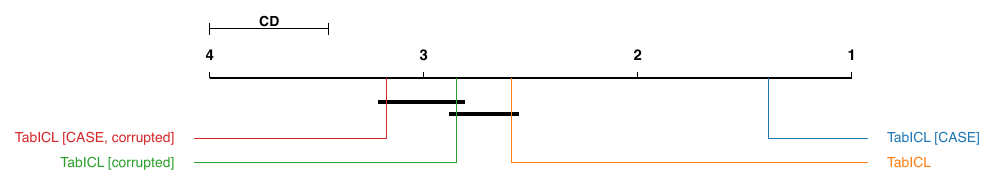}
    \caption{Critical difference diagram under semantic corruptions: Across CARTE and TextTab benchmarks ($\alpha = 0.05$), standard TabICL [CASE] achieves a significantly superior performance rank compared to all non-semantic or corrupted configurations.}
    \label{fig:cd_corrupted}
\end{figure*}

\subsection{Importance of Semantic Signals}
\label{sec:appendix-semantics}

\subsubsection{Impact of Semantic Corruption}

To further evaluate whether CASE leverages true semantic understanding or merely exploits statistical patterns, we conduct a \emph{semantic corruption test}. 

\custompar{Experimental Setup}
We designed a deterministic character-substitution preprocessor that applies a consistent substitution cipher across all column headers and string/textual/categorical cell values. Numerical values and overall row/column structure remain identical. This completely strips away real-world domain semantics (e.g., column headers like \texttt{"Age"} or cell values like \texttt{"Doctor"}) while preserving identical categorical frequencies. We evaluate two baseline pipelines on CARTE and TextTab benchmarks under both original and corrupted regimes:
\begin{enumerate}[leftmargin=*]
    \item \textbf{TabICL}: Pure non-semantic baseline utilizing TF-IDF encoding.
    \item \textbf{TabICL [CASE]}: Combined pipeline integrating CASE-12B semantic embeddings with TF-IDF.
\end{enumerate}

\custompar{Statistical Impact and Analysis}
Results are depicted in Table~\ref{tab:semantic-corruption-ablation}.
Three key insights emerge regarding when and why CASE provides performance gains:
\begin{itemize}[leftmargin=*]
    \item \textbf{Performance Degradation upon Corruption:} When semantics are corrupted, the mean rank of \mbox{TabICL [CASE, corrupted]} drops sharply from $1.3$ to $3.0$, returning to a level statistically indistinguishable from the baseline TabICL model ($2.3$).
    \item \textbf{Statistical Superiority of Uncorrupted CASE:} The CD-diagram in Fig.~\ref{fig:cd_corrupted} confirms that the superiority of CASE leveraging semantic context over scrabled text is statistically significant, placing it in its own distinct rank group.
    \item \textbf{Robustness of Non-Semantic Baselines:} Standard \mbox{TabICL} exhibits virtually identical performance between uncorrupted ($78.3\%$ Acc, $73.9\%\ R^2$) and corrupted ($77.7\%$ Acc, $73.4\%\ R^2$) datasets on CARTE, as expected for TF-IDF representations.
\end{itemize}

\custompar{Conclusion -- when CASE Helps}
These results confirm that CASE's performance gains are driven by \emph{}{genuine semantic content}. When semantic information is present, CASE lifts performance substantially. When semantic information is absent or destroyed, CASE gracefully degrades to a standard non-semantic tabular modeling baseline performance.

\subsubsection{Performance Correlation with a Semantic Proxy}
Finally, to investigate in detail how predictive performance depends on the semantic density of a dataset, we investigate performance improvements of TabICL with CASE over bare TabICL over a semantic density proxy across the TabArena as well as text-heavy benchmarks (CARTE and TextTab).

\custompar{Semantic Density Proxy}
To quantify \emph{how much} of a dataset's predictive information is carried by
genuine natural language -- as opposed to numerical or categoricals --
we define a scalar \emph{semantic density} proxy $\rho \in [0, 1]$ for each table.
Intuitively, $\rho$ measures the share of the table's total column-wise information
that resides in natural-language text columns.
We first partition the (target-excluded) feature columns into string and numeric
columns. For each string column $c$, we tokenize all cell values and compute two
quantities: its token-frequency Shannon entropy $H_c = -\sum_t p_t \log_2 p_t$
(in bits), which captures the raw information content of the column, and a
\emph{textiness} weight $w_c \in [0,1]$, the fraction of tokens that are alphabetic
and of length $\geq 3$. The latter is a lightweight, language-agnostic gate that
down-weights columns dominated by identifiers, single-symbol codes, or short
abbreviations, which are high-entropy yet carry no
exploitable linguistic semantics. For each numeric column $c$ we compute a binned
Shannon entropy $H_c$ over $32$ equal-width histogram bins. The semantic density is
then the textiness-weighted string entropy normalized by the total column entropy,
\begin{equation}
    \rho \;=\;
    \frac{\sum_{c \in \mathcal{S}} w_c \, H_c}
         {\sum_{c \in \mathcal{S}} H_c \;+\; \sum_{c \in \mathcal{N}} H_c},
\end{equation}
where $\mathcal{S}$ and $\mathcal{N}$ denote the sets of string and numeric
columns, respectively. By construction, numeric columns only enter the denominator
and can therefore only \emph{dilute} the score, while non-linguistic string columns
are suppressed through $w_c$; both known confounds are thus excluded by design. A
value of $\rho \to 1$ indicates a table whose information is almost entirely
natural-language text, whereas $\rho \to 0$ corresponds to a purely numeric or
code-based table. To bound computation on large tables, we subsample at most
$5{,}000$ rows per dataset before computing the statistic.

\begin{figure*}[t]
    \centering
    \includegraphics[width=0.7\linewidth]{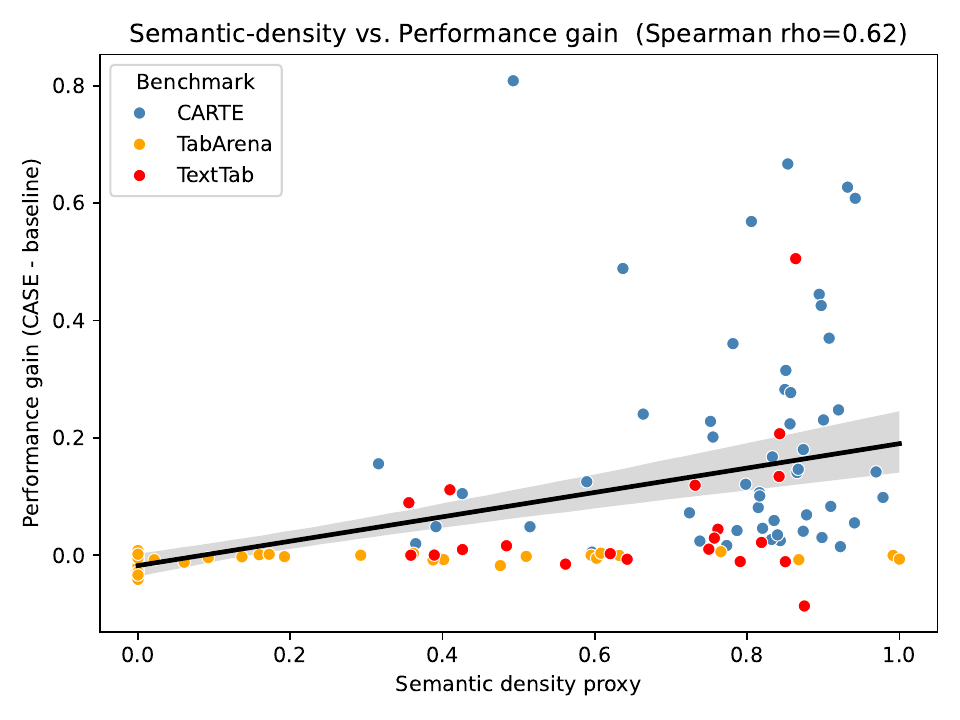}
    \caption{Performance gains of using CASE over a vanilla TabICL baseline depending on the semantic density proxy of the underlying task across the investigated benchmarks.}
    \label{fig:app-semantic-density-proxy}
\end{figure*}

The results are depicted in Figure~\ref{fig:app-semantic-density-proxy}.
First, we observe that the proxy nicely distinguishes TabArena versus text-heavy benchmarks (CARTE and TextTab), which contain tasks with much higher semantic density.
Next, we confirm that performance gains of using CASE vs vanilla TabICL strongly correlates with the semantic density (Spearman correlation of 62\%) confirming that CASE successfully leverages semantic content in those predictive tasks.

%%%%%%%
%%% COLUMN ORDER PERMUTATIONS
%%%%%%%

\begin{table}[t]
\centering\footnotesize
\caption{Robustness to column order permutation: Performance across 5 random column ordering seeds using CASE-12B with TabICLv2. Ranks and metrics show negligible variance.}
\label{tab:column-permutation-ablation}
\begin{tabular}{l S[table-format=1.1] S[table-format=1.1] S[table-format=2.1] S[table-format=2.1] S[table-format=1.1] S[table-format=2.1] S[table-format=2.1]}
\toprule
& \multicolumn{1}{c}{All} & \multicolumn{3}{c}{CARTE} & \multicolumn{3}{c}{TextTab} \\
\cmidrule(lr){2-2} \cmidrule(lr){3-5} \cmidrule(lr){6-8}
Column Seed & {Rank} & {Rank} & {Acc} & {R2} & {Rank} & {Acc} & {R2} \\
\midrule
Seed 1      & 2.3 & 2.4 & 81.3 & 77.4 & 2.2 & 86.1 & 68.9 \\
Seed 2     & 2.5 & 2.7 & 80.8 & 77.5 & 2.0 & 86.2 & 68.9 \\
Seed 3    & 2.7 & 2.6 & 81.1 & 77.5 & 3.0 & 85.8 & 67.8 \\
Seed 4   & 2.5 & 2.5 & 81.0 & 77.6 & 2.6 & 85.9 & 67.8 \\
Seed 5  & 2.6 & 2.6 & 80.7 & 77.5 & 2.6 & 85.9 & 66.6 \\
\midrule
\textbf{Mean $\pm$ Std} & $\mathbf{2.5 \pm 0.1}$ & $\mathbf{2.6 \pm 0.1}$ & $\mathbf{81.0 \pm 0.2}$ & $\mathbf{77.5 \pm 0.1}$ & $\mathbf{2.5 \pm 0.4}$ & $\mathbf{86.0 \pm 0.2}$ & $\mathbf{68.0 \pm 1.0}$ \\
\bottomrule
\end{tabular}
\end{table}

\begin{figure}[t]
    \centering
    \includegraphics[width=\linewidth]{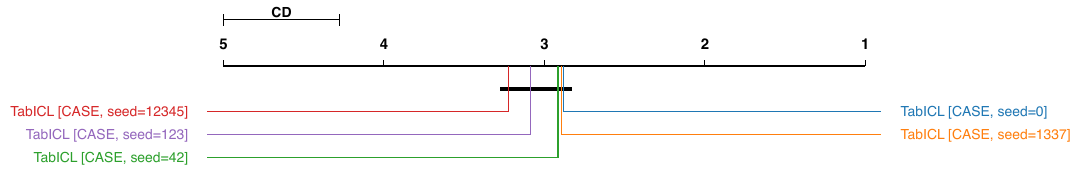}
    \caption{Critical difference diagrams for CARTE and TextTab benchmarks using different seeds for column permutations.}
    \label{fig:cd_col_perm}
\end{figure}

\subsection{Impact of Column Order Permutation}
\label{sec:appendix-column-permutation}

CASE is neither row nor column permutation equivariant. 
While we have previously investigated the robustness with respect to context sampling and row order, we now want to evaluate whether the order of input columns impacts the downstream embedding quality.
To this end. we conduct an ablation study over 5 distinct column ordering seeds using the CASE-12B model.

\custompar{Statistical Equivalence via CD Diagrams}
We performed a Friedman test followed by a post-hoc Nemenyi test ($\alpha = 0.05$) across CARTE and TextTab benchmarks to construct Critical Difference (CD) diagrams for the different column ordering seeds. In the resulting CD diagram (Fig~\ref{fig:cd_col_perm}), all 5 seed variants are connected by a single, continuous critical difference bar. While some fluctuations exist, this confirms that there is \emph{no statistically significant difference} in overall performance ranks across any of the column permutations.

\custompar{Performance Stability}
As shown in Table~\ref{tab:column-permutation-ablation}, performance across both CARTE and TextTab benchmarks remains exceptionally tight across all seeds. For instance, the average classification accuracy on CARTE fluctuates within a minimal margin of $81.0\% \pm 0.2\%$, while the overall average rank varies by only $\pm 0.1$. 

These results demonstrate that while CASE is architecturally non-equivariant to column permutations, its learned semantic representations are robust with respect to column ordering.

%%%%%%%
%%% RANDOM CONTEXT SELECTION
%%%%%%%
\newpage

\begin{table}
\centering\footnotesize
\caption{Robustness to random context selection: Performance across 5 random context sampling seeds using CASE-12B with the Skrub preprocessor on TabICL. Ranks and metrics show negligible variance.}
\label{tab:context-selection-ablation}
\begin{tabular}{l S[table-format=1.1] S[table-format=1.1] S[table-format=2.1] S[table-format=2.1] S[table-format=1.1] S[table-format=2.1] S[table-format=2.1]}
\toprule
& \multicolumn{1}{c}{All} & \multicolumn{3}{c}{CARTE} & \multicolumn{3}{c}{TextTab} \\
\cmidrule(lr){2-2} \cmidrule(lr){3-5} \cmidrule(lr){6-8}
Context Seed & {Rank} & {Rank} & {Acc} & {R2} & {Rank} & {Acc} & {R2} \\
\midrule
Seed 1      & 2.7 & 2.8 & 80.8 & 77.4 & 2.4 & 85.9 & 68.3 \\
Seed 2     & 2.3 & 2.5 & 80.8 & 77.5 & 1.9 & 86.2 & 68.9 \\
Seed 3    & 2.5 & 2.4 & 80.7 & 77.6 & 3.0 & 85.6 & 67.9 \\
Seed 4  & 2.9 & 2.8 & 80.5 & 77.3 & 3.0 & 85.9 & 67.7 \\
Seed 5  & 2.4 & 2.4 & 80.9 & 77.6 & 2.2 & 85.9 & 68.6 \\
\midrule
\textbf{Mean $\pm$ Std} & $\mathbf{2.5 \pm 0.2}$ & $\mathbf{2.6 \pm 0.2}$ & $\mathbf{80.7 \pm 0.2}$ & $\mathbf{77.5 \pm 0.1}$ & $\mathbf{2.5 \pm 0.5}$ & $\mathbf{85.9 \pm 0.2}$ & $\mathbf{68.3 \pm 0.5}$ \\
\bottomrule
\end{tabular}
\end{table}

\begin{figure*}
    \centering
    \includegraphics[width=\linewidth]{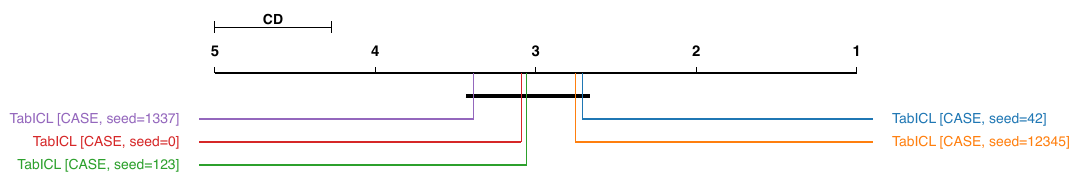}
    \caption{Critical difference diagrams for CARTE and TextTab benchmarks using different seeds for random context selection.}
    \label{fig:cd_context_perm}
\end{figure*}

\subsection{Impact of Random Context Selection}
\label{sec:appendix-context-permutation}

CASE relies on a randomly selected context to prefill the KV cache. To evaluate whether CASE exhibits instability due to stochastic context selection, we evaluated the CASE-12B model across 5 independent evaluation runs. Each run draws a completely distinct random sample of context rows with a randomized permutation of row order.

\custompar{Statistical Equivalence via CD Diagrams}
To test whether random context selection leads to statistically significant performance variations, we performed a Friedman test followed by a post-hoc Nemenyi test ($\alpha = 0.05$) across all benchmark datasets. The resulting Critical Difference (CD) diagram connects all 5 context selection seeds under a single continuous critical difference bar. This statistically confirms that varying the random seed for context row selection produces no significant difference in overall model performance.

\custompar{Performance Stability}
As detailed in Table~\ref{tab:context-selection-ablation}, performance across all context selection seeds remains remarkably stable. The overall average rank stays tightly bounded around $2.5 \pm 0.2$, with average accuracy on CARTE showing a tight standard deviation of only $0.2\%$ ($80.7\% \pm 0.2\%$). Similarly, regression performance ($R^2$) varies by a maximum of $0.3\%$ across all runs on CARTE and TextTab.

These results indicate that CASE effectively distills the broader underlying semantic distribution of the tabular dataset rather than over-indexing on specific, high-leverage context examples or arbitrary row arrangements.

%%%%%%%
% Sensitivity to PCA Dimension Reduction
%%%%%%%

\newpage

\begin{table}
\centering
\footnotesize
\caption{Sensitivity to PCA embedding dimensions: Increasing PCA components monotonically improves representation expressiveness while $k=32$ serves as a strong default.}
\label{tab:pca-components-ablation}
\begin{tabular}{l S[table-format=1.1] S[table-format=1.1] S[table-format=2.1] S[table-format=2.1] S[table-format=1.1] S[table-format=2.1] S[table-format=2.1]}
\toprule
& \multicolumn{1}{c}{All} & \multicolumn{3}{c}{CARTE} & \multicolumn{3}{c}{TextTab} \\
\cmidrule(lr){2-2} \cmidrule(lr){3-5} \cmidrule(lr){6-8}
Configuration & {Rank} & {Rank} & {Acc (\%)} & {$R^2$ (\%)} & {Rank} & {Acc (\%)} & {$R^2$ (\%)} \\
\midrule
TabICL [CASE, $k=256$] & \bfseries 1.9 & \bfseries 1.6 & \bfseries 81.3 & \bfseries 77.8 & 2.4           & \bfseries 86.3 & 69.0           \\
TabICL [CASE, $k=128$] & \bfseries 1.9 & 1.8           & 81.0           & 77.7           & \bfseries 1.9 & 86.2           & \bfseries 69.6 \\
TabICL [CASE, $k=64$]  & 2.2           & 2.1           & 80.7           & 77.7           & 2.5           & 86.1           & \bfseries 69.6 \\
TabICL [CASE, $k=32$]  & 2.8           & 2.8           & 80.8           & 77.5           & 2.6           & 86.2           & 68.9           \\
TabICL [CASE, $k=16$]  & 3.4           & 3.4           & 80.7           & 77.2           & 3.4           & 85.8           & 68.2           \\
\bottomrule
\end{tabular}
\end{table}

\begin{figure*}[t]
    \centering
    \includegraphics[width=\linewidth]{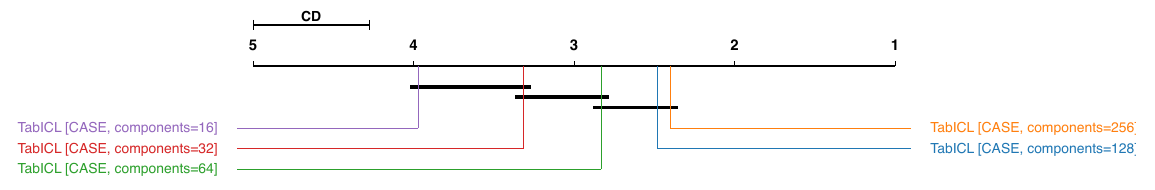}
    \caption{Critical difference diagram across PCA dimensions. Higher PCA dimensions ($k=128, 256$) yield statistically significant rank improvements across CARTE and TextTab benchmarks ($\alpha = 0.05$).}
    \label{fig:cd_pca_components}
\end{figure*}

\subsection{Sensitivity to PCA Dimension Reduction}
\label{sec:appendix-pca-components}

To evaluate how embedding dimensionality affects downstream prediction quality, we conducted an ablation study over the number of Principal Component Analysis (PCA) components ($k \in \{16, 32, 64, 128, 256\}$) used to compress CASE representations.

\custompar{Statistical Ranking via CD Diagrams}
We evaluated ranking performance across CARTE and TextTab benchmarks using a Friedman test followed by a post-hoc Nemenyi test ($\alpha = 0.05$). As illustrated in the Critical Difference (CD) diagram (Fig.~\ref{fig:cd_pca_components}), increasing the target dimension yields progressive performance gains:
\begin{itemize}[leftmargin=*]
    \item \textbf{High-Dimensional Regimes ($k \in \{128, 256\}$):} Models using $128$ and $256$ components achieve the best overall ranks ($1.9$), significantly outperforming lower-dimensional settings ($k \in \{16, 32\}$).
    \item \textbf{Standard Default ($k = 32$):} Our default setting of $32$ components forms a statistically equivalent clique with $k=16$ and $k=64$.
\end{itemize}

\custompar{Performance Trade-offs \& Analysis}
As detailed in Table~\ref{tab:pca-components-ablation}, performance scales predictably with dimensionality: Classification accuracy remains stable across settings (varying by only $\sim 0.6\%$ across $k=16$ to $256$), whereas regression metrics ($R^2$) benefit noticeably from higher dimensions ($+0.6\%$ on CARTE, $+1.4\%$ on TextTab moving from $k=16$ to $k=128$).

%%%%%%%%%%%%%%
%%%%% COMPUTATIONAL EFFICIENCY AND RUNTIME
%%%%%%%%%%%%%%

\newpage
\section{Computational Efficiency and Runtime Analysis} \label{app:efficiency}

While integrating a Tabular Language Model (TLM) into tabular learning pipelines introduces additional computational overhead compared to shallow feature extraction, CASE is explicitly designed to maintain a practical trade-off between rich semantic context and runtime throughput.

\custompar{Architectural Optimizations} 
The efficiency of CASE relies on two key architectural principles:
\begin{enumerate}[leftmargin=*,label=\arabic*.]
    \item \textbf{KV-Cache Context Priming:} Background row representations and table-level semantics are processed and cached during \texttt{fit()}. At prediction time, context rows do not need to be re-encoded, avoiding redundant forward passes.
    \item \textbf{Generation-Free Inference:} Unlike standard LLM applications that rely on autoregressive token sampling -- the dominant computational bottleneck in LLM inference -- CASE extracts target representations from the final hidden states in a single, fast forward pass without invoking the language modeling head.
\end{enumerate}

\custompar{Online Prediction vs. Offline Fitting Trade-off}
A key advantage of CASE is the isolation of heavy neural compute to the offline fitting phase, \texttt{fit()}. At inference time, \texttt{predict()}, CASE projects row embeddings into a fixed $k=32$ PCA feature budget regardless of the number of text columns ($C$) in the input table. 
In contrast, standard TF-IDF pipelines typically extract up to 32 n-gram components \emph{per text column}, expanding the downstream feature space to $32 \times C$. For tables with numerous textual attributes, this dramatic expansion increases the attention complexity of downstream tabular architectures (such as TabICL).

\begin{figure*}[t]
\centering
\begin{subfigure}[t]{0.49\textwidth}
    \centering
    \includegraphics[width=\linewidth]{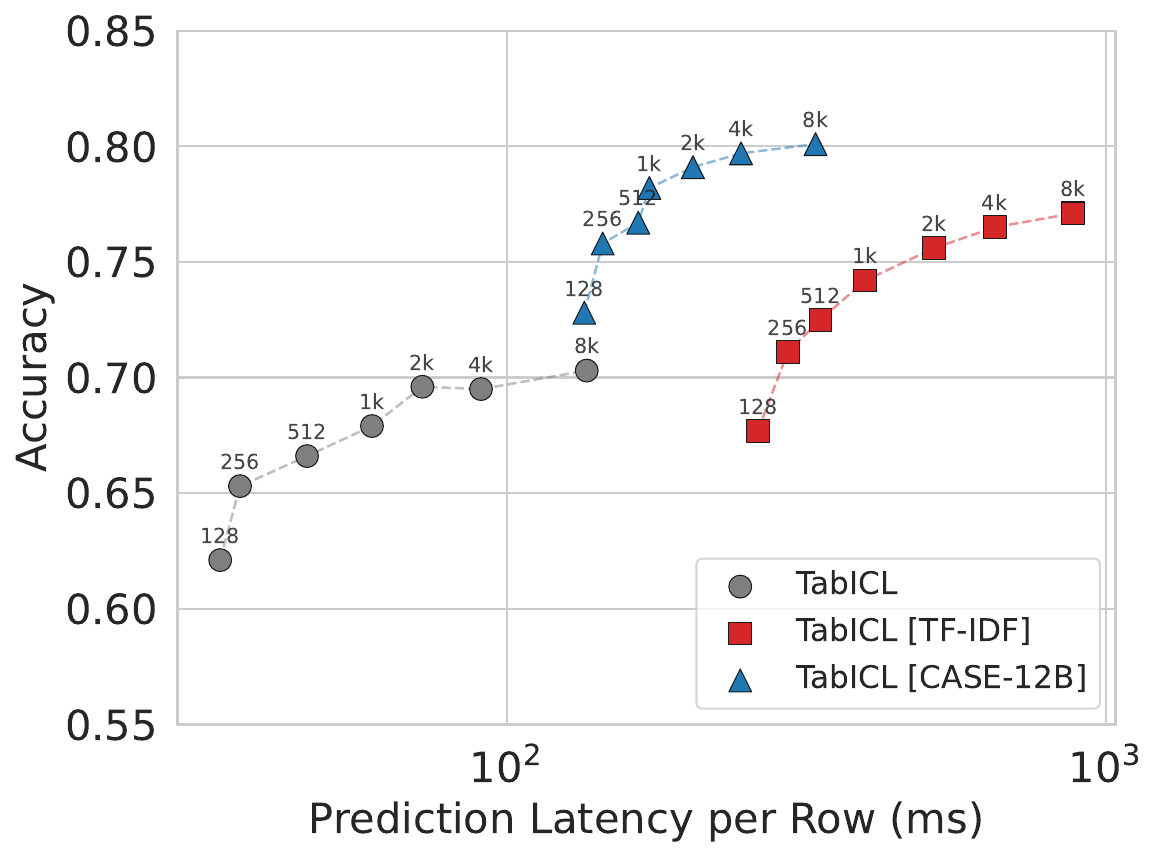} 
    \caption{Online Prediction Latency per Row (ms): CASE achieves superior predictive quality while maintaining faster online inference than TF-IDF due to a constant downstream feature dimension budget.}
    \label{fig:pareto_latency}
\end{subfigure}\hfill
\begin{subfigure}[t]{0.49\textwidth}
    \centering
    \includegraphics[width=\linewidth]{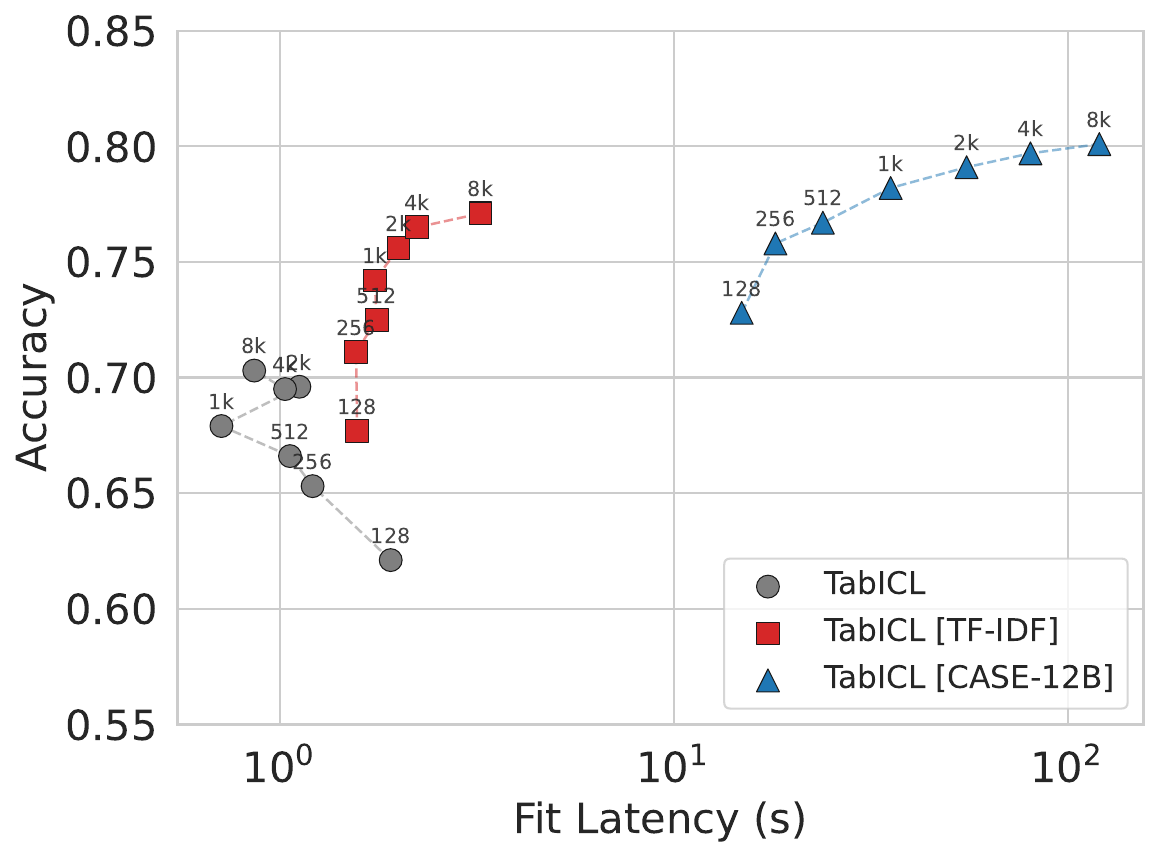}
    \caption{Offline Fit Latency (s): Heavy backbone compute is isolated to training time, scaling predictably with sample size.}
    \label{fig:pareto_fit}
\end{subfigure}
\caption{Predictive Quality vs. Computational Cost Trade-offs on CARTE across subsampled training set sizes $N \in \{128, 256, 512, 1024, 2048, 4096, 8192\}$. Pareto frontiers demonstrate that CASE offers highly competitive inference latency while improving predictive performance over non-semantic baselines.}
\label{fig:pareto}
\end{figure*}

\custompar{Empirical Latency Benchmarks}
To quantitatively evaluate throughput and predictive quality, we benchmarked TabICL, TabICL [TF-IDF], and TabICL [CASE-12B] across subsampled training set sizes $N \in \{128, 256, 512, 1024, 2048, 4096, 8192\}$ on 11 classification tasks from the CARTE benchmark while keeping the test evaluation sets fixed. All experiments were conducted on a single NVIDIA H100 GPU. As illustrated in Figure~\ref{fig:pareto}, CASE consistently defines the Pareto-optimal frontier for the prediction latency.

%%%%%%%%%%%%%
%%%%% DATA CONTAMINATION
%%%%%%%%%%%%

\section{Data Contamination and Leakage Analysis}\label{app:contamination}
As CASE leverages the pretrained Gemma 3 backbone and utilizes the T4 dataset \cite{tabula8b} for continued pretraining, we proactively address the potential for data contamination:. 
Given that our evaluation benchmarks -- CARTE, TextTab, and TabArena -- are open-source, there is a risk that samples from these datasets were present in the LLM's original pretraining corpus or the T4 collection which we used for continued pretraining of the TLM.

\custompar{T4 Contamination Study} Related work \cite{contexttab} conducted a systematic search for overlap between the T4 training set and the CARTE benchmark. By performing string similarity matching on unique combinations of column names and cell values, and found no evidence of CARTE samples within the T4 dataset. This suggests that the performance gains observed via CASE are not the result of direct leakage during our continued pretraining phase.

\custompar{Backbone Memorization} While the exact composition of the Gemma 3 pretraining data is proprietary, our empirical results suggest that the model has not memorized these benchmarks in a way that bypasses tabular reasoning. As shown in our various ablations, the model's predictive accuracy is highly sensitive to the number of context rows provided in the KV cache. If the model were relying on memorized labels, we would expect a high baseline performance with zero or minimal context; instead, we observe a clear monotonic improvement as more in-context information is provided. This behavior indicates that the model is actively conditioning on the provided tabular distribution at inference time rather than retrieving stored labels from its weights.

%%%%%%%%%%%
%%% EVAL ON STRABLE
%%%%%%%%%%%

\section{Evaluation on STRABLE Benchmark}
\label{sec:appendix_strable}

To further assess the generalization of CASE on string-heavy tabular datasets, we extend our empirical evaluation to the recently introduced STRABLE benchmark suite~\cite{strable}. STRABLE evaluates tabular learners across a diverse collection of classification and regression tasks characterized by high-cardinality unstructured text and complex string features.

We evaluate TabICLv2 augmented with CASE-12B against established tabular baselines, specialized string-tabular models (such as ConTextTab), and automated machine learning frameworks (AutoGluon). 

\begin{table}
\centering
\footnotesize
\caption{Performance comparison on the STRABLE benchmark. Models are ordered by overall mean rank across all classification and regression tasks. Best performance is highlighted in \textbf{bold}.}
\label{tab:strable_results}
%\vspace{0.5em}
\begin{tabular}{lccc}
\toprule
\textbf{Model Name} & \textbf{Overall Rank} ($\downarrow$) & \textbf{Accuracy (\%)} ($\uparrow$) & \textbf{R$^2$ (\%)} ($\uparrow$) \\
\midrule
\textbf{TabICLv2 [CASE-12B]} & \textbf{2.0} & \textbf{76.3} & \textbf{72.8} \\
AutoGluon                     & 2.3          & 75.7          & 69.9          \\
ConTextTab                    & 2.8          & 74.6          & 69.2          \\
RealMLP                       & 3.3          & 71.9          & 68.0          \\
TabPFN-2.6                    & 5.0          & 65.3          & 54.8          \\
TabICLv2                      & 5.5          & 65.1          & 49.1          \\
Naive                         & 6.7          & 56.0          & --4.7         \\
\bottomrule
\end{tabular}
\end{table}

\begin{figure}
\centering
\includegraphics[trim={30mm 15mm 0 0}, clip, width=\linewidth]{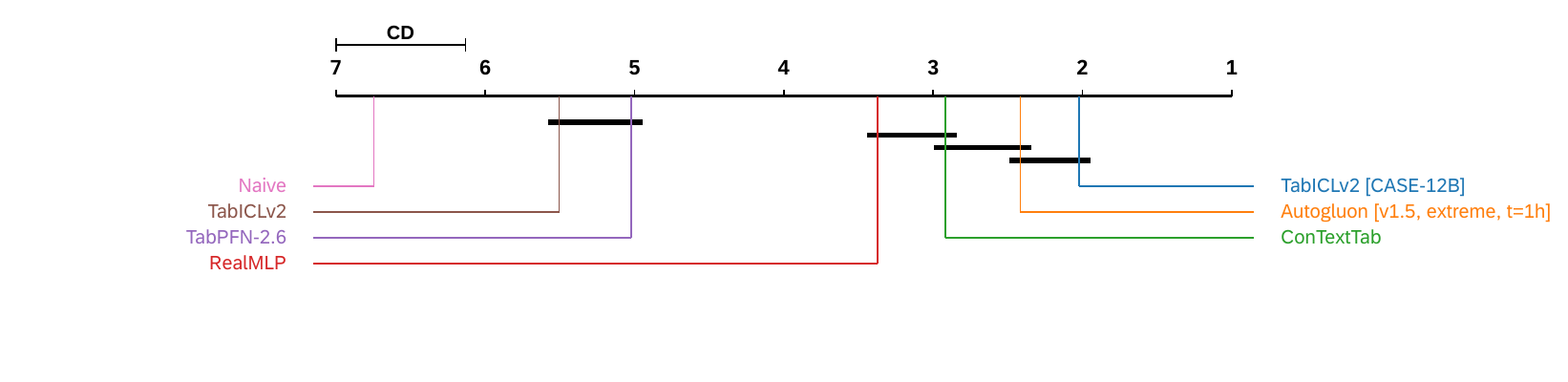} 
\caption{Critical difference diagram across the STRABLE benchmark.}
\label{fig:cd-strable}
\end{figure}

CASE-12B achieves the top overall rank across the benchmark, outperforming domain-specific string pipelines like ConTextTab as well as strong AutoML ensembles like AutoGluon. Integrating CASE-12B into TabICLv2 provides a large performance boost over standard TabICLv2, lifting classification accuracy from $65.1\%$ to $76.3\%$ ($+11.2\%$) and regression $R^2$ from $49.1\%$ to $72.8\%$ ($+23.7\%$). Critical Difference (CD) diagram analysis ($\alpha = 0.05$) confirms that TabICLv2 [CASE-12B] is statistically superior to all evaluated baselines, with AutoGluon being the only competitor within the critical distance threshold, as shown in Figure~\ref{fig:cd-strable}.

\end{document}